\documentclass[10pt,letterpaper]{article}
\usepackage[top=0.85in,left=2.75in,footskip=0.75in]{geometry}

\usepackage{nameref}

\usepackage{amsmath,amssymb}

\usepackage{changepage}

\usepackage{textcomp,marvosym}

\usepackage{cite}

\usepackage{nameref,hyperref}

\usepackage[nopatch=eqnum]{microtype}
\DisableLigatures[f]{encoding = *, family = * }

\usepackage[table]{xcolor}

\usepackage{array}
\usepackage{spotcolor}

\newcolumntype{+}{!{\vrule width 2pt}}

\newlength\savedwidth

\raggedright
\usepackage[aboveskip=1pt,labelfont=bf,labelsep=period,justification=raggedright,singlelinecheck=off]{caption}

\makeatletter
\renewcommand{\@biblabel}[1]{\quad#1.}
\makeatother

\usepackage{lastpage,fancyhdr,graphicx}
\usepackage{epstopdf}
\fancyheadoffset[L]{2.25in}
\fancyfootoffset[L]{2.25in}
\NewSpotColorSpace{PANTONE}
\AddSpotColor{PANTONE} {PANTONE3015C} {PANTONE\SpotSpace 3015\SpotSpace C} {1 0.3 0 0.2}
\SetPageColorSpace{PANTONE}
\definecolor{accessblue}{cmyk}{1, 0.3, 0, 0.2}
\definecolor{greycolor}{cmyk}{0,0,0,.8}

\usepackage{cite}
\usepackage{amsmath,amssymb,amsfonts}
\usepackage{algorithmic}
\usepackage{graphicx}
\usepackage{textcomp}
\usepackage{pgfplots}
\usepackage{multirow}
\usepackage{comment}
\usepackage{ragged2e}
\usepackage{wrapfig}

\usepackage{pgfplots}
\pgfplotsset{compat=1.18} 

\usepackage{bm}
\makeatletter
\AtBeginDocument{\DeclareMathVersion{bold}
\SetSymbolFont{operators}{bold}{T1}{times}{b}{n}
\SetMathAlphabet{\mathrm}{bold}{T1}{times}{b}{n}
\SetMathAlphabet{\mathit}{bold}{T1}{times}{b}{it}
\SetMathAlphabet{\mathbf}{bold}{T1}{times}{b}{n}
\SetMathAlphabet{\mathtt}{bold}{OT1}{pcr}{b}{n}
\SetSymbolFont{symbols}{bold}{OMS}{cmsy}{b}{n}
\renewcommand\boldmath{\@nomath\boldmath\mathversion{bold}}}
\makeatother

\usepackage{adjustbox}
\usepackage{url}
\usepackage{textcomp}
\usepackage{booktabs}
\usepackage{makecell}
\usepackage{multirow}
\usepackage{tikz}
\usepackage{pgfplots}
\usetikzlibrary{calc}
\usepackage {color, graphicx}
\usetikzlibrary{decorations.pathmorphing}
\usetikzlibrary {patterns,shapes,shadows,decorations.pathreplacing}
\usepackage{amssymb}
\usetikzlibrary{arrows,shadows,positioning}
\usetikzlibrary{arrows.meta}
\usetikzlibrary{snakes}
\usepackage{rotating}
\usepackage{ifthen}
\usepackage{amssymb}
\usepackage{pifont}
\usetikzlibrary{shapes.symbols}
\usepackage{graphicx,subfig}
\usepackage{color, colortbl}	
\usepackage{lineno}
\usepgfplotslibrary{groupplots}
\pgfplotsset{compat=1.18}

\usepackage{forest}

\usetikzlibrary{positioning} 

\usetikzlibrary{shapes.geometric, arrows, positioning}

\tikzstyle{startstop} = [rectangle, rounded corners, minimum width=3cm, minimum height=1cm, text centered, draw=black, fill=gray!20]
\tikzstyle{process} = [rectangle, minimum width=3cm, minimum height=1cm, text centered, draw=black, fill=blue!20]
\tikzstyle{decision} = [diamond, minimum width=3cm, minimum height=1cm, text centered, draw=black, fill=green!20]
\tikzstyle{arrow} = [thick,->,>=stealth]

\usepackage{longtable}
\usepackage[x11names]{xcolor}  

\def\BibTeX{{\rm B\kern-.05em{\sc i\kern-.025em b}\kern-.08em
    T\kern-.1667em\lower.7ex\hbox{E}\kern-.125emX}}

\newcommand{\etal}{\textit{et~al.}}

\newcommand\modelTaskOneLongo{Multimodal Large Language Model}
\newcommand\modelTaskOneCurto{MLLM}

\newcommand\propAppCurto{MLLMsent}

\newcommand\modelTaskTwoLongo{Large Language Model}
\newcommand\modelTaskTwoCurto{LLM}
\newcommand\VLMosLongo{MiniGPT-4 (Open Source)}
\newcommand\VLMos{GPT (OS)}
\newcommand\VLMoaiLongo{GPT-4o mini (OpenAI)}
\newcommand\VLMoai{GPT (OAI)}
\newcommand\VLMdsLongo{DeepSeek-VL2-Tiny}
\newcommand\VLMds{DeepSeek}

\newcommand\TaskOne{\textit{Task 1}}
\newcommand\TaskTwo{\textit{Task 2}}
\newcommand\TaskTwoA{\textit{Task~2\textsubscript{a}}}
\newcommand\TaskTwoB{\textit{Task~2\textsubscript{b}}}

\newcommand{\paramProb}[3]{$\langle{#1},{#2}_{{#3}}\rangle$}

\newcommand\captioning{image description} 
\newcommand\Captioning{Image Description} 
\newcommand\imgDescrip{image description}

\newcommand{\Dataset}{PerceptSent}
\newcommand{\DeepSent}{DeepSent}
\newcommand{\evaluator}{evaluator}

\newcommand{\probVersion}{P}
\newcommand\vectorSent{\mathbf{s}}
\newcommand\vectorSentElement{s}

\usetikzlibrary{matrix}
\tikzset{ 
    table/.style={
        matrix of nodes,
        row sep=-\pgflinewidth,
        column sep=-\pgflinewidth,
        font=\scriptsize,
        nodes={
            align=center
        },
        minimum height=1.5em,
        text depth=0.5ex,
        text height=2ex,
        nodes in empty cells,
        column 1/.style={
            nodes={text width=5em}
        },
    }
}

\begin{document}
\vspace*{0.2in}

\begin{flushleft}
{\Large
\textbf{Multimodal LLMs see sentiment} 
}
\newline
\\
Neemias B. da Silva\textsuperscript{1},
John Harrison\textsuperscript{1},
Rodrigo Minetto\textsuperscript{1},
Myriam R. Delgado\textsuperscript{1},
Bogdan T. Nassu\textsuperscript{1},
Thiago H. Silva\textsuperscript{2*}
\\
\bigskip
\textbf{1} Universidade Tecnológica Federal do Paraná, Curitiba, Brazil
\\
\textbf{2} University of Toronto, Toronto, Canada
\\
\bigskip

%
%





* th.silva@utoronto.ca (TS)

\end{flushleft}


%
\section*{Abstract}
Understanding how visual content conveys sentiment is increasingly important in a digital landscape dominated by imagery. However, sentiment perception depends on complex scene-level semantics, making this a challenging task for computational models. This paper examines how Multimodal Large Language Models (MLLMs) perform sentiment analysis in images through a systematic, evaluation-driven study encompassing three perspectives: (i) direct sentiment classification from images using MLLMs; (ii) sentiment analysis on MLLM-generated descriptions using pre-trained LLMs; and (iii) fine-tuning these LLMs on sentiment-labeled descriptions to assess performance and generalization. Experiments on a recent benchmark show that a two-stage MLLM description-mediated pipeline can substantially improve prediction accuracy under several evaluation settings, particularly when the LLM component is fine-tuned. Across different agreement thresholds and sentiment granularities, the strongest configurations of this pipeline outperform lexicon-, CNN-, and Transformer-based baselines in our benchmark by up to 30.9\%, 64.8\%, and 42.4\%, respectively. In cross-dataset evaluation, the proposed pipeline — without training or fine-tuning on the target dataset — still surpasses the best in-domain baseline by over 8\%. Overall, the study provides a comprehensive assessment of MLLM description-mediated sentiment analysis, clarifying the conditions under which it is effective, the scenarios in which it fails, and its comparison with traditional vision-based approaches, while also providing a reproducible benchmark resource for future research.


\section*{Introduction}\label{sec:introduction}

Visual sentiment analysis, or image sentiment analysis, seeks to automatically predict the emotions conveyed by visual content as perceived by human observers. 
This problem has significant implications for computational social systems. It is rooted in the need to better understand collective behavior, public sentiment, and societal trends in digital environments, as social platforms increasingly rely on visual communication~\cite{VLMforSentAnalysis2023,zisad2021integrated}; and images often express emotions more powerfully than text alone~\cite{chandrasekaran2022visual}. 

Although recent studies have highlighted the importance of visual attributes in images, such as color, texture, and shape~\cite{ortis2020survey}, for sentiment prediction, the inherent complexity and richness typically provided by images, where even subtle elements can alter emotional perception, make this a challenging problem. Lopes~\etal~\cite{PerceptSent} report that the integration of human-annotated textual tags --- describing perceived elements such as the presence of nature, violence, or lack of maintenance --- can substantially improve visual sentiment analysis performance, with $F$-score gains of up to 35\%. However, automatically extracting such subjective, perception-level cues remains difficult, as it requires models that go beyond object recognition toward scene interpretation and affective reasoning.

Recent multimodal Large Language Models (MLLMs) offer a new possibility for this problem: instead of learning image sentiment directly from visual features, they can verbalize what they ``see'' and reason about it in language form, externalizing latent perceptual cues. This shifts the task from traditional vision-only classification to a two-stage MLLM description-mediated pipeline, which can potentially be useful in various tasks \cite{kil2024mllm,zhang2024mm,cheng2025evaluating}. Among the potential advantages of this approach, we highlight its ability to express
scene understanding through explicit textual descriptions, enabling the use of mature text-based sentiment models and improving interpretability, which is increasingly recognized as important in multimodal AI systems \cite{10.1145/3746027.3755591,ansari2025role,10.1145/3723005,ZHANG2025112812,ansari2025optimized}. Moreover, such pipelines can better capture contextual and scene-level semantic cues that may be difficult to infer from visual features alone.  However, despite the increasing use of such pipelines in practice, there has been little systematic analysis of when description-mediated reasoning helps,  when it fails, and how its performance compares with previous baselines in visual sentiment analysis.

We introduce a systematic evaluation framework designed to answer four research questions that probe how and why MLLM-based pipelines behave: (Q1) How effective are MLLMs for direct sentiment classification from raw images? (Q2) Does transforming images into textual descriptions and analyzing the resulting descriptions with pre-trained LLMs provide measurable benefits? (Q3) How does fully fine-tuning LLMs on sentiment-labeled descriptions affect performance and generalization? (Q4) How does the best-performing MLLM-based pipeline compare with traditional vision-based and text-based approaches in both in-domain and in a cross-dataset setting?

Our goal is to provide a systematic, reproducible, and behavior-oriented evaluation of description-mediated pipelines based on MLLMs in visual sentiment analysis, identifying gains, trade-offs, and relevant patterns. Rather than exhaustively comparing all multimodal sentiment architectures, this work focuses on isolating and systematically evaluating the behavior of MLLM-based direct visual reasoning and description-mediated reasoning for image-level sentiment analysis under a unified protocol. 

In response to these research questions, our contributions are as follows:

\begin{itemize}
    
\item We conduct a systematic and multi-setting evaluation of three MLLMs — MiniGPT-4~\cite{zhu2024minigpt}, GPT-4o mini~\cite{openai2024gpt4ocard}, and DeepSeek-VL2-Tiny~\cite{wu2024deepseek} — characterizing their behavior under different sentiment granularities and annotator-agreement thresholds; here we perform direct sentiment classification from raw images, i.e., isolating MLLM visual reasoning without description mediation.

\item We analyze MLLM description-mediated pipelines in which images are first translated into textual descriptions by MLLMs and then classified using LLM-based text classifiers (BART~\cite{lewis2020bart,williams2018broad}, ModernBERT~\cite{warner2024modernBert}, and LLAMA-3~\cite{dubey2024llama3}), and we investigate the conditions under which this strategy improves prediction relative to direct image-level classification.

\item We assess the impact of fully fine-tuning the text classifiers on MLLM-generated descriptions, evaluating effects on accuracy, robustness, and cross-dataset transfer behavior.

\item We benchmark the MLLM pipeline against traditional CNN-, Transformer-, and lexicon-based models, to contextualize their strengths and limitations across in-domain and cross-dataset scenarios. 

\item We release all generated descriptions, fine-tuned models, and experimental scripts to facilitate reproducibility and future benchmarking studies\footnote{https://github.com/neemiasbsilva/multimodal-LLMs-see-sentiment}.

\end{itemize}

This paper is structured as follows. Section~\nameref{sec:Related} reviews existing approaches to textual, visual, and multimodal sentiment analysis. Section~\nameref{sec:dataset} introduces the \Dataset\ and DeepSent benchmarks and formalizes the sentiment analysis task. Section~\nameref{sec:methodology} details the proposed framework, including the MLLM and LLM configurations. Section~\nameref{sec:experiments} presents the experimental setup and analyzes the obtained results. Finally, Section~\nameref{sec:conclusions} summarizes the main findings and discusses directions for future research.

\section*{Related work}\label{sec:Related}

\textbf{Textual Sentiment Analysis.} The field of textual sentiment analysis has evolved significantly from early lexicon-based approaches~\cite{jurek2015improved,jha2018novel} to sophisticated deep learning models. Traditional methods rely on sentiment dictionaries and rule-based systems~\cite{bernabe2020context,viegas2020exploiting}, whereas modern approaches leverage neural architectures, including BERT-based models~\cite{10478509} and graph neural networks~\cite{LIANG2022107643,WU2022107736}.
A recent work~\cite{Yin_Zhong_2024} shows that transformers capture aspect-level sentiment through double-view graph representations.
The success of textual analysis has motivated its use in multimodal contexts, such as video transcription~\cite{stappen2021sentiment}.
However, challenges remain in platforms like Instagram, where visual content dominates and textual context is sparse~\cite{Ji2016}.

\textbf{Visual Sentiment Analysis.} Deep learning has revolutionized visual sentiment analysis, with convolutional neural networks (CNNs) playing a pivotal role. An early work~\cite{you2015robust} introduces probability sampling to reduce label noise, while~\cite{cnn1} combines visual features with web-mined sentiment concepts. Subsequent advances incorporate attention mechanisms~\cite{SONG2018218} and scene semantics~\cite{outdoorsent2020} to improve performance. The importance of human perception in visual sentiment is highlighted in~\cite{PerceptSent} --- the work shows that incorporating evaluator judgments significantly improves model accuracy. Despite these advances, purely visual pipelines still face difficulties when sentiment depends on subjective or context-dependent cues.

\textbf{Multimodal Approaches.} Combining visual and textual modalities has emerged as a powerful paradigm for sentiment analysis~\cite{PANDEY2024111206}. In~\cite{das2023image}, the authors demonstrate the effectiveness of late fusion strategies, whereas interaction networks capture cross-modal dependencies in~\cite{zhu2022multimodal}. Recent work introduced architectures such as the multi-stage perception model~\cite{pan2024multi} and the dual-perspective fusion network~\cite{wang2023dual}, which better align visual and textual representations. These methods show that joint modeling of modalities can address the limits of unimodal approaches, especially for ambiguous or context-dependent sentiment expressions. These advances set the stage for a new paradigm, where vision-language understanding is embedded into large language models.

\textbf{Multimodal LLMs.}  The emergence of 
MLLMs~\cite{cheng2025evaluating,kil2024mllm,zhang2024mm,liu2025unveiling} has expanded the possibilities for multimodal sentiment analysis in computational social systems. While traditional LLMs like GPT-4o and LLAMA-2~\cite{zhao2024exploring} have advanced sentiment analysis~\cite{zhang2023sentimentanalysiseralarge}, their text-only nature faces limitations in data scarcity~\cite{villalobos2024position} and modality constraints~\cite{hong2024only}. To address these limitations, MLLMs integrate visual encoders, enabling joint processing of images and text. Examples include DeepSeek-VL~\cite{lu2024deepseek}, DeepSeek-VL2~\cite{wu2024deepseek}, MiniGPT-4~\cite{zhu2024minigpt}, and GPT-4o mini~\cite{openai2024gpt4ocard}. Despite differences in scale and architecture, these models share the goal of enhancing LLMs with perceptual capabilities for tasks like visual sentiment analysis.   Studies~\cite{xiao2025exploring,huang2023language} suggest that MLLMs exhibit human-like perception, but their behavior in affective and sentiment-centric tasks remains only partially understood.

While these multimodal approaches and MLLMs achieve strong performance through joint cross-modal modeling, they differ from the focus of this work, which investigates how MLLMs perform sentiment reasoning either directly from images or through description-mediated pipelines under a unified evaluation framework. Therefore, this paper examines the behavior of MLLMs in visual sentiment analysis, comparing different model families and training strategies to evaluate when image-to-text transformations improve performance, introduce errors, or diverge from conventional visual baselines.
 
\section*{Datasets and problem formulation}\label{sec:dataset}

In the present study, we use two datasets: {\Dataset}~\cite{PerceptSent} and {\DeepSent}~\cite{you2015robust}. The former is a publicly available\footnote {\url {https://github.com/ceslop84/perceptsent.}} collection of 5,000 images, most (92.7\%) of which depict outdoor scenes. The images have been sourced from Instagram, Flickr, and NYC311\footnote{https://portal.311.nyc.gov.}. The dataset has been fully annotated by human evaluators using mainly Amazon Mechanical Turk\footnote{https://www.mturk.com.} (AMT). In \DeepSent, a set of 1,269 images collected from X (formerly Twitter), covering indoor and outdoor scenes, was annotated through the AMT crowdsourcing platform.

To ensure comparability with prior work and support reproducibility, we adopt the problem formulation for sentiment analysis proposed by Lopes \emph{et al.}~\cite{PerceptSent}. Denoted here as {\paramProb{\sigma_l}{P}{C}}, we have  $C$ as the number of sentiment categories considered in a problem instance $P_C$, and $l$ as a threshold for filtering out divergent {\evaluator}s opinions, allowing us to define different types of sentiment dominance for filter $\sigma_l$. 

Formally, let $\vectorSent= (\vectorSentElement_1,...,\vectorSentElement_C)$ represent the \textit{sentiment} vector for a given image, where $\vectorSentElement_c$ denotes the number of votes assigned to sentiment category $c$ by the {\evaluator}s. Since each image is evaluated by $E$ {\evaluator}s, we have $\sum_{c=1}^C \vectorSentElement_c = E$. 

The {\Dataset} dataset considers five sentiment categories ($C = 5$) --- positive, slightly positive, neutral, slightly negative, and negative --- with a nearly balanced distribution. Each image is independently annotated by five {\evaluator}s ($E = 5$). 
Specifically, in vector~$\vectorSent$, $\vectorSentElement_1$ represents the total votes for a positive sentiment, $\vectorSentElement_2$ for slightly positive, $\vectorSentElement_3$ for neutral, $\vectorSentElement_4$ for slightly negative, and $\vectorSentElement_5$ for negative. The dataset comprises 25,000 sentiment evaluations (5,000 images $\times$ 5 evaluations per image).

The {\DeepSent} dataset considers only two sentiment categories ($C = 2$) --- positive and negative. The distribution is slightly imbalanced, with a higher proportion of positive instances. Each image is also independently annotated by five {\evaluator}s ($E = 5$), with 6,345 sentiment evaluations (1,269 images $\times$ 5 evaluations per image).

\begin{figure}[!htb]

\captionsetup[subfloat]{justification=centering}

    \centering

    \subfloat[PerceptSent]{%
        \includegraphics[width=0.49\linewidth]{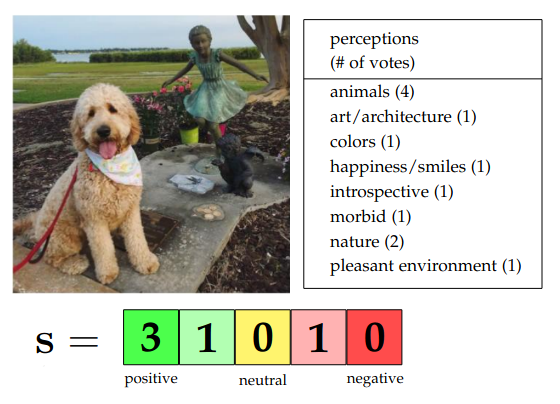}
    }
    \hfill
    \subfloat[DeepSent]{%
        \includegraphics[width=0.49\linewidth]{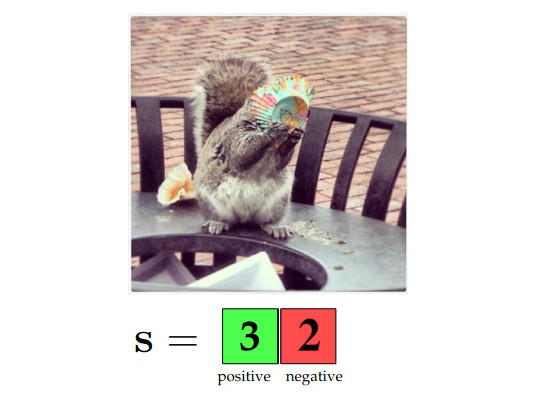}
    }

    \caption{Image samples and associated data from (a) PerceptSent and (b) DeepSent datasets, with the corresponding sentiment polarity vector ($\vectorSent$) assigned by human evaluators. This figure is original to this work. Image (a) is derived from the PerceptSent dataset~\cite{Cesar}, distributed under the Creative Commons Attribution 4.0 International (CC BY 4.0) license. Image (b) is derived from the DeepSent GitHub repository~\cite{CamposGit}, distributed under the MIT License, which permits reuse, modification, and redistribution with attribution.}

    \label{fig:sample_dataset_here}
\end{figure}

Fig.~\ref{fig:sample_dataset_here} presents one sample image from PerceptSent and one from DeepSent, as well as the data associated with each sample. In PerceptSent, human-provided perception tags capture scene attributes that help explain how annotators arrived at a sentiment judgment. In contrast, for DeepSent, the absence of perception annotations highlights a setting where sentiment must be inferred directly from visual content alone. This distinction is relevant to our study, as it allows us to evaluate description-mediated reasoning both in the presence and absence of explicit perceptual cues.

A \textbf{dominant} sentiment or category in the \textit{sentiment} vector $\vectorSent$ is then defined as:
\begin{equation}
\label{eq:dominant}
    c_{\sigma_l}^* (\vectorSent) =  \{arg\max_c \; \vectorSent = (\vectorSentElement_1,...,\vectorSentElement_C)\;|\;\max (\vectorSent) \ge l \text{ \small{in} } \sigma_l\}
\end{equation}
where $\max (\vectorSent)$ is the maximum value among all elements of $\vectorSent$ for a particular image, i.e., the number of votes of the most voted category in the image; and $\sigma_l$ is set to define a minimum number of votes for a specific target label in the pool of five {\evaluator}s addressed in the dataset. For example, for $\sigma_3$, only images where at least three {\evaluator}s agreed on a sentiment category are included. We consider the following values for 
$\sigma_l$: $\sigma_3$ (\textbf{simple dominance}) and $\sigma_5$ (\textbf{absolute dominance}).  

Furthermore, as in \cite{PerceptSent}, we also explore variations in the number $C$ of output target classes to evaluate the impact of different levels of separability in sentiment classification. This enables us to examine how MLLM description-mediated pipelines behave under tasks of varying difficulty, from fine-grained sentiment distinctions to binary polarity. In this work, we consider three variations (denoted as $\probVersion_C$), ranging from the most challenging case ($C=5$) to the simplest ($C=2$):

\begin{itemize}

\item $\probVersion_5$ considers all five sentiment categories as originally assigned by {\evaluator}s in PerceptSent --- positive, slightly positive, neutral, slightly negative, and negative, with the  \textit{sentiment}  vector  $\vectorSent$ given by  $\vectorSent= (\vectorSentElement_1,...,\vectorSentElement_5)$.

\item $\probVersion_3$ merges in PerceptSent slightly positive with positive evaluations and slightly negative with negative, $\vectorSent= (\vectorSentElement'_1,\vectorSentElement'_2,\vectorSentElement'_3)$, with $\vectorSentElement'_1 = \vectorSentElement_1+\vectorSentElement_2$, $\vectorSentElement'_2 = \vectorSentElement_3$,  and $\vectorSentElement'_3 = \vectorSentElement_4+\vectorSentElement_5$.

\item $\probVersion_{2}$  for a binary (positive \emph{versus} negative) classification problem, as defined for DeepSent. 
\end{itemize}

Note that the higher the $l$ level in $\sigma_l$, the smaller the subset resulting from filtering the entire dataset, due to a stronger consensus requirement.  As with $l$ in $\sigma_l$, the higher the number of classes $C$ in a problem $P_C$, the smaller the cardinality $|\{{P_C}\}|$ of the resulting subset of images (for {\Dataset}, $|\{P_3\}| > |\{P_5\}|$). For example, an image excluded from the subset of images $\{P_5\}$ due to the threshold comparison could still be included in $\{P_3\}$ as its sentiment values increase ($\vectorSentElement'_1 \geq \vectorSentElement_1$, $\vectorSentElement'_1 \geq \vectorSentElement_2$, $\vectorSentElement'_3 \geq \vectorSentElement_4$, and $\vectorSentElement'_3 \geq \vectorSentElement_5$, for each given $\sigma_l$). Then, in the experiments, we have 4 subsets from {\Dataset}: for $\sigma_3$,  $|\{P_3\}| = 4,506 $ and $|\{P_5\}| = 3,566$; and for $\sigma_5$,  $|\{P_3\}| = 1,680$ and $|\{P_5\}| = 446$.  Similarly, in {\DeepSent}, we obtain $|\{P_2\}| = 1,269$ for $\sigma_3$, and $|\{P_2\}| = 882$ for $\sigma_5$.

These filtered subsets are consistently used across all experiments, enabling controlled comparisons between models and configurations.

\section*{Methodology}\label{sec:methodology}

\begin{figure}[!htb]
\centering
\includegraphics[scale=0.45]{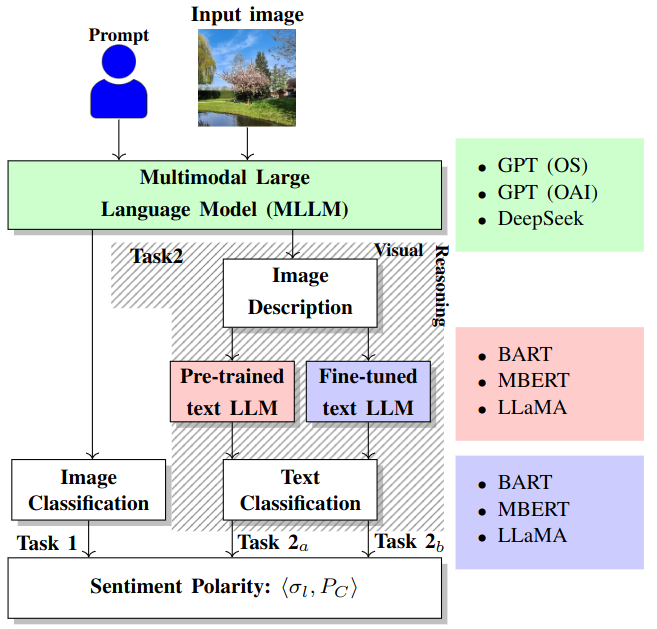}
\caption{Architecture diagram of the Multimodal Large Language Model framework for sentiment analysis. This figure is original to this work; images used within the figure are from the PerceptSent dataset~\cite{Cesar}, licensed under CC BY 4.0.}
\label{fig:overviewProposal}
\end{figure}

This section describes the evaluation framework we adopt. As shown in Fig.~\ref{fig:overviewProposal}, different models can be used for two main tasks. In \textit{Task}~$1$, a {\modelTaskOneLongo}, {\modelTaskOneCurto} for short, directly classifies, following the prompt instructions, the sentiment associated with each image to be examined. In \textit{Task}~$2$, we adopt an MLLM description-mediated pipeline (hatched area in Fig.~\ref{fig:overviewProposal}), in which the {\modelTaskOneCurto}s first generate a textual description for each image, and this description is subsequently analyzed by a text-only {\modelTaskTwoLongo} ({\modelTaskTwoCurto}) for sentiment classification. Throughout the paper, the term MLLM description-mediated pipeline refers to this two-stage process of image description followed by text-based sentiment reasoning. This second task can be addressed using pre-trained models (\textit{Task} $2_a$), or by fine-tuning the text classifier on sentiment-labeled descriptions (\textit{Task}~$2_b$). In the following sections, we outline the models used for image classification,  {\captioning}, and text-based sentiment classification, as well as the experimental setup employed in our study.

\subsection*{Multimodal large language models ({\modelTaskOneCurto}s)}

We evaluate three models for the {\modelTaskOneCurto} component (highlighted in green in Fig.\ref{fig:overviewProposal}), which serves as the core of \textit{Task}~1 and the first stage of \textit{Task}~2. The compared models are the open-source MiniGPT-4~\cite{zhu2024minigpt}, the proprietary GPT-4o mini~\cite{openai2024gpt4ocard}, and the open-weights~\VLMdsLongo{} \cite{wu2024deepseek}. 

We intentionally selected models representing different capability levels and deployment paradigms. MiniGPT-4 represents an open-source vision-language architecture for local experimentation and reproducibility; GPT-4o mini represents a commercially optimized multimodal reasoning system; and DeepSeek-VL2-Tiny represents a lightweight open-weights alternative for constrained computational environments. This diversity enables a broader evaluation of how different MLLM configurations affect description quality and downstream sentiment reasoning.

{\textbf{\VLMosLongo}}: {\VLMos} for short, is an open-source vision-language model introduced by \cite{zhu2024minigpt}. Claiming that the technical details behind OpenAI's GPT-4 continue to remain
undisclosed, the authors of {\VLMos} proposed a model built upon Vicuna~\cite{zheng2023judgingllmasajudgemtbenchchatbot}, a large language model (LLM) derived from LLAMA~\cite{touvron2023llama}. In terms of visual perception, the authors employ the same pre-trained vision components of~BLIP-2~\cite{li2023blip} that consist of a ViT-G/14 from EVA-CLIP~\cite{fang2023eva} and a Q-Former network, which can transform users' queries based on the image. {\VLMos} adds a projection layer to align encoded visual features with the Vicuna language model, while freezing the remaining components. Although architecturally simple, the model is reported to perform a wide range of multimodal tasks, including detailed image description. The instance used in our experiments contains approximately 7 billion parameters (Vicuna-based).

{\textbf{\VLMoaiLongo}}: {\VLMoai}, for short, is a proprietary, closed-source, closed-weights multimodal model developed by OpenAI, offered as a lighter, more affordable version of GPT-4o. The exact number of internal parameters or architecture details is also undisclosed. Despite its smaller size compared to other models being offered as services, GPT-4o mini has achieved competitive performance on several benchmarks. In our experiments, GPT-4o mini is accessed as a paid cloud API service, in contrast to the locally executed open and open-weights models.

{\textbf{\VLMdsLongo}}: {\VLMds} for short, is part of the DeepSeek-VL2 series, and introduces advanced Mixture-of-Experts (MoE) Vision-Language Models (VLMs) designed for multimodal understanding. Specifically, the tiny version we consider features around 3 billion parameters, of which 1 billion are activated during inference, allowing evaluation under more constrained GPU conditions.

To promote fair comparison across models, we have experimented with the aforementioned MLLMs using various prompt formulations, including those generated by the models themselves. We selected the most effective prompt for each task, balancing clarity, consistency, and model response accuracy:
\begin{itemize}
    \item {\TaskOne} (prompt for Image Classification):     ``\textit{Analyze this image, and classify it as $\{\mathcal{L}\}$ sentiments, do not describe the image, and select only one class}.'' Here, $\{\mathcal{L}\}$ is the set of sentiment labels (or categories) in each classification setting $P_C$ (e.g., when the set of labels in $P_3$ is \{positive, neutral, negative\}, we have $C =|\{\mathcal{L}\}|= 3$. 
    
   \item {\TaskTwo} (prompt for  {\Captioning}): ``\textit{Describe this image in details.}''
\end{itemize}

In the visual reasoning scheme proposed in {\TaskTwo}, each {\modelTaskOneCurto} (the models within the green box in Fig. \ref{fig:overviewProposal})  provides a textual {\imgDescrip} for every image. The description is then passed to a text-only {\modelTaskTwoCurto}, which classifies the sentiment. The objective is to classify each textual description into a predefined sentiment category based on its linguistic features.

\subsection*{Text-only large language models ({\modelTaskTwoCurto}s)}

In this study, we compare different LLMs to classify the sentiment of generated image descriptions. 
The models were selected based on their ability to process contextual and affective cues in text, enabling analysis of how architecture, scale, and training strategy influence description-mediated sentiment reasoning. The following sections detail the two approaches used for {\TaskTwo}: application of the text-only LLMs with their pre-trained weights (\TaskTwoA) and fine-tuning the weights to PerceptSent images (\TaskTwoB).

We consider three architectures:  BART-Large-MNLI~\cite{lewis2020bart,williams2018broad}, ModernBERT~\cite{warner2024modernBert}, and LLAMA-3~\cite{dubey2024llama3} to examine how model size, context capacity, and architectural design affect sentiment classification performance on image descriptions.

{\textbf{BART-LARGE-MNLI}~\cite{lewis2020bart,williams2018broad}:} 
BART, for short, is a sequence-to-sequence transformer model pre-trained as a denoising autoencoder, capable of capturing complex contextual relationships in the text. Its original fine-tuning, performed on the Multi-Genre Natural Language Inference (MultiNLI) dataset~\cite{williams2018broad}, endows BART with a natural language inference capability, which we further adapt through a second fine-tuning stage (\TaskTwoB) for sentiment classification over image descriptions.

{\textbf{ModernBERT}~\cite{warner2024modernBert}:} MBERT, for short, is an encoder-only transformer model pre-trained on an extensive corpus of 2 trillion tokens. It supports a native sequence length of 8,192 tokens, enabling it to efficiently handle long-context inputs. The architecture incorporates several modern optimizations, including rotary positional embeddings (RoPE), GeGLU activation functions, bias-free linear layers, alternating global/local attention mechanisms, unpadding techniques, and \emph{flash attention}. These design choices balance computational efficiency with strong downstream performance.

{\textbf{LLAMA-3}~\cite{dubey2024llama3}:}  LLAMA-3 is a decoder-only, autoregressive Transformer that follows the design principles of LLAMA-2 while introducing improvements in scalability, efficiency, and alignment. 

In our study, these LLMs are used in their pre-trained form or fine-tuned to perform sentiment classification on image descriptions. Both settings involve a weight adjustment procedure (minimal in the case of pre-trained models and more significant for fine-tuned ones).

{\textbf{Pre-trained LLMs}:} \label{sec:pre-trained}
 For {\TaskTwoA}, we consider LLMs with very few modifications. BART and MBERT are extended with a single task-specific linear layer mapping the output to the number of target sentiment classes, i.e., the number $C$ in each problem $P_C$. All the remaining weights of those pre-trained models are kept fixed. For LLAMA, we use a prompt engineering approach called few-shot learning, in which a subset of random samples (image descriptions taken from the training set) is included directly in the prompt for classification. In our experiments, we tested subsets ranging from 5 to 15 samples. Smaller subsets proved insufficient, while the 15-sample subset emerged as the best alternative, albeit slightly more computationally expensive.
 
{\textbf{Fine-tuned LLMs}:} \label{sec:fine-tuning} For {\TaskTwoB}, LLMs are fine-tuned to enable each model to adapt its internal representations (weights) to the specific language patterns and sentiment cues found in textual {\imgDescrip}s of PerceptSent images. The fine-tuning process proposed for BART and  MBERT starts from the pre-trained models (extended with the final linear mapping to the $C$ classes), and adjusts the whole pipeline weights via supervised learning on our labeled image-description data. In this work, we perform LLAMA fine-tuning through qLORA (Quantized Low-Rank Adapters), which applies low-rank adapters to pre-trained models, significantly reducing their memory footprint while preserving high performance. 

When adjusting the weights in {\TaskTwoA} and {\TaskTwoB}, the training loop iteratively updates model parameters over several epochs and incorporates early stopping based on validation F1-score improvements to prevent overfitting. The instances used as LLM inputs are textual descriptions provided by MLLMs. Such descriptions are framed as text generation tasks using sentiment-aware prompts adapted to the number of classes in each configuration $P_C$, where $C \in \{2,3,5\}$. Each prompt instructs the model to determine the sentiment of a given description by selecting from a predefined set of labels. For example, in the three-class setting $P_3$, the prompt is ``What is the sentiment of this description? Please choose an answer from \{``Positive": 2, ``Negative": 0, ``Neutral": 1\}''.  Then they are tokenized and go through data-loaders with dynamic batching and shuffling for robust training. 

This design guides the model to generate sentiment predictions consistent with the class granularity defined by $P_C$ and the agreement threshold $l$ in filter $\sigma_l$. Class imbalance in the sentiment labels is addressed by weighting the cross-entropy loss inversely proportional to class frequencies. Aiming to improve analyses' robustness, in both tasks, $2_a$ and $2_b$, a stratified 5-fold cross-validation scheme assesses model performance.  
    
\subsection*{Experimental setup }~\label{sec:setup} 

The experimental framework is implemented in Python 3.10.12 using the PyTorch v2.7.1, Hugging Face Transformers v4.14.0, trl v0.18.1, Sklearn v1.7.0, and BitsandBytes v0.45.3 libraries. The experiments have been carried out on a machine equipped with an Intel(R) Xeon(R) Silver 4316 CPU @2.30GHz, 256GB of RAM, and two GPUs: an NVIDIA RTX 4000 Ada Generation (20GB VRAM) and an NVIDIA RTX A6000 (48GB VRAM). Considering all configurations (see Table~\ref{tab:resultsSentVLMs}), training times for BART range from [1.2 - 17.9] hours, for MBERT from [1.0 - 12.9] hours, and for LLAMA fine-tuning range from [1.8 - 29.3] hours, with LLAMA consistently exhibiting the highest computational cost. The differences in training time and model scale may also highlight practical trade-offs between predictive performance and computational efficiency. While larger MLLMs and fine-tuned LLMs generally achieve stronger results, lightweight alternatives such as DeepSeek-VL2-Tiny remain attractive for resource-constrained environments, suggesting that scalability and deployment cost should be considered when selecting models for real-world applications.

For {\VLMos}, image classification (\TaskOne) and {\imgDescrip} (first step in \TaskTwo)  are  governed by two tuned hyperparameters: \textit{temperature} $=0.1$ and \textit{beam search} $=1$. The former controls the randomness of word selection and influences the variability of the output; the latter enables the model to consider multiple candidate sequences and select the most appropriate one. For {\VLMoai}, both tasks rely on the built-in default generation parameters, with \textit{temperature} fixed at 1.0, \textit{max token length} set to 300 to constrain output size, and all other settings left at their API defaults. This configuration provides a balanced trade-off between creativity and consistency, leveraging the model's internal heuristics for token selection without additional constraints or penalties. For DeepSeek,  the tasks are controlled by five hyperparameters (left as default): \textit{max\_new\_tokens} (512), \textit{repetition\_penalty} (1.1), \textit{do\_sample} (set as true), \textit{temperature} (0.1), and \textit{top\_p} (0.9). \textit{Max\_new\_tokens} sets an upper bound on the length of the generated sequence, preventing overly verbose outputs; \textit{repetition\_penalty} discourages the model from recycling the same phrases; \textit{do\_sample} enables stochastic sampling rather than greedy decoding; \textit{temperature} modulates randomness in token selection; and \textit{top\_p} (nucleus sampling) restricts the sampling pool to the smallest set of tokens whose cumulative probability exceeds 0.9, balancing coherence and novelty.

For the LLMs, we adopt the following general configuration unless specified otherwise. We use the AdamW optimizer -- a learning rate of $2 \times 10^{-3}$ for a {\TaskTwoA} and $2 \times 10^{-5}$ for {\TaskTwoB}, both with a weight decay of $0.01$. Models are trained for up to $100$ epochs. We employ an early stopping mechanism with a patience of up to 25 epochs. This strategy was adopted to reduce overfitting and improve training stability during fine-tuning, particularly in configurations with smaller training subsets.

\section*{Results}~\label{sec:experiments}

In this section, we present results from the experiments performed to address the four research questions outlined earlier. Section ``Direct sentiment classification using {\modelTaskOneCurto}s'' addresses \textbf{(Q1)} by evaluating the performance of different \modelTaskOneCurto{} models on direct image-based sentiment classification (\TaskOne). Section ``Pre-trained LLMs for classifying MLLM image descriptions'' responds to \textbf{(Q2)} by investigating whether converting images into textual descriptions using \modelTaskOneCurto{} and classifying them with pre-trained LLMs (\TaskTwoA) provides gains over direct visual prediction. Section ``Fine-tuning LLMs for classifying MLLM image descriptions'' explores \textbf{(Q3)}, examining the impact of fully fine-tuning LLMs for sentiment classification using labeled \modelTaskOneCurto{}-generated textual descriptions (\TaskTwoB). To investigate \textbf{(Q4)}, we compare the strongest-performing MLLM-based (description-mediated) configuration with conventional approaches, including Lexicon-, CNN-, and Transformer-based models. This comparison spans both the \textit{PerceptSent} dataset (Section ``{\propAppCurto} \textit{vs} language and image-based baselines'') and a cross-dataset generalization to \textit{DeepSent} (Section ``Cross-dataset evaluation of MLLMsent'').

Table~\ref{tab:resultsSentVLMs} groups the main results for {\TaskOne} and {\TaskTwo} on PerceptSent and is discussed throughout the following subsections. As defined in Section ``Datasets and Problem Formulation'', the parameter $l \in \{3, 5\}$ in the filter $\sigma_l$ specifies the minimum agreement level among {\evaluator}s, ranging from simple consensus ($\sigma_3$) to absolute consensus ($\sigma_5$), whereas $P_C$, $C \in \{3, 5\}$, specifies the number of sentiment classes for each sentiment scale in the problem setup. $C=2$ is considered only in the DeepSent analysis (see Section ``Cross-dataset evaluation of MLLMsent'').

\begin{table*}[!htb]
\begin{adjustwidth}{-2.25in}{0in}
\footnotesize
\caption{$F$-score results ($\pm$ confidence intervals of 95\%) across different MLLM and LLM combinations for \textbf{Task 1} (direct sentiment classification from images with MLLMs) and \textbf{Task 2} (sentiment classification based on MLLM-generated {\imgDescrip}s processed by a text-only LLM), evaluated under different sentiment polarities and annotator agreement levels for the PerceptSent dataset. In \textbf{Task~2$_a$}, pre-trained LLMs perform text classification, while in \textbf{Task~2$_b$}, their fine-tuned counterparts are considered.
Highlighted cells indicate the best average performance within each task: green for \textbf{Task 1}, pink for \textbf{Task~2$_a$}, blue for \textbf{Task~2$_b$}. }
\setlength{\tabcolsep}{7pt}
\renewcommand{\arraystretch}{1.2}
\centering

\begin{tabular}{c|c|c|lclclc}
\hline
\multirow{4}{*}{\textbf{Problem}} & \multirow{4}{*}{\textbf{\modelTaskOneCurto}} & & \multicolumn{6}{c}{\textbf{Classification of \textbf{\modelTaskOneCurto}'s Scene Textual Descriptions}} \\ \cline{4-9}  & & \textbf{Image} & \multicolumn{2}{c|}{\textbf{BART}}  & \multicolumn{2}{c|}{\textbf{MBERT}} & \multicolumn{2}{c}{\textbf{LLAMA}} \\ &  & \textbf{Classification}  & \multicolumn{1}{c}{\textbf{Task $2_a$}} & \multicolumn{1}{c|}{\textbf{Task $2_b$}} & \multicolumn{1}{c}{\textbf{Task $2_a$}}      & \multicolumn{1}{c|}{\textbf{Task $2_b$}} & \multicolumn{1}{c}{\textbf{Task $2_a$}} & \textbf{Task $2_b$} \\ &  & \textbf{Task 1} & \multicolumn{1}{c}{\scriptsize pre-trained} & \multicolumn{1}{c|}{\scriptsize fine-tuned} & \scriptsize pre-trained & \multicolumn{1}{c|}{\scriptsize fine-tuned} & \scriptsize pre-trained & \scriptsize fine-tuned \\ \hline \multirow{3}{*}{\normalsize $\langle \sigma_3,P_5 \rangle$} & \VLMos & --- & $36.5$ \scriptsize $\pm 2.6$  & $48.0$ \scriptsize $\pm 2.6$ & $43.4$ \scriptsize $\pm 1.1$  & $51.2$ \scriptsize $\pm 1.6$ & $24.4$ \scriptsize $\pm 3.7$ & $49.5$ \scriptsize $\pm 2.9$ \\ & \VLMoai                                      & \cellcolor[HTML]{90EE90}$\textbf{44.5}$\scriptsize $\pm 3.1$ & $40.1$ \scriptsize $\pm 3.0$ & $56.1$ \scriptsize $\pm 2.5$  & \cellcolor[HTML]{FFCCCC}\textbf{47.9} \scriptsize $\pm 2.3$  & \cellcolor[HTML]{CCCCFF}\textbf{58.4} \scriptsize $\pm 4.0$ & $33.3$ \scriptsize $\pm 6.3$ & $56.9$ \scriptsize $\pm 2.9$ \\ & DeepSeek & $26.6$ \scriptsize $\pm 1.9$  & $34.6$ \scriptsize $\pm 3.0$ & $52.2$ \scriptsize $\pm 2.5$ & $38.2$ \scriptsize $\pm 1.7$  & $50.5$ \scriptsize $\pm 1.4$ &  $25.3$ \scriptsize $\pm 9.0$ & $51.0$ \scriptsize $\pm 2.4$ \\ \hline \multirow{3}{*}{\normalsize $\langle \sigma_3,P_3 \rangle$} & \VLMos & ---                                                                                                & $59.4$ \scriptsize $\pm 1.7$                & $68.7$ \scriptsize $\pm 2.1$                                               & $62.1$ \scriptsize $\pm 8.9$  & $69.8$ \scriptsize $\pm 1.7$                & $30.3$ \scriptsize $\pm 11.1$ & $70.9$ \scriptsize $\pm 2.6$ \\
                                                                                                   & \VLMoai                                      & \cellcolor[HTML]{90EE90}
\textbf{61.2} \scriptsize $\pm 2.9$  & $64.9$ \scriptsize $\pm 2.4$                & $76.0$ \scriptsize $\pm 0.7$                                               & \cellcolor[HTML]{FFCCCC}\textbf{66.6} \scriptsize $\pm 6.1$  & \cellcolor[HTML]{CCCCFF}\textbf{77.5} \scriptsize $\pm 0.9$                &  $46.6$ \scriptsize $\pm 14.0$ & $76.7$ \scriptsize $\pm1.7$  \\ & DeepSeek   & $44.0$ \scriptsize $\pm 0.2$                                    & $57.2$ \scriptsize $\pm 1.5$                & $71.1$ \scriptsize $\pm 1.2$                                               & $60.6$ \scriptsize $\pm 5.2$  & $71.1$ \scriptsize $\pm 1.6$                & $46.6$ \scriptsize $\pm 7.5$ & $72.9$ \scriptsize $\pm 3.5$ \\ \hline
\multirow{3}{*}{\normalsize $\langle \sigma_5,P_5 \rangle$} & \VLMos                                       & ---                                                                                                & $54.1$ \scriptsize $\pm 4.5$               & $80.1$ \scriptsize $\pm 3.3$                                               & $65.4$ \scriptsize $\pm 3.5$ & $75.7$ \scriptsize $\pm 4.8$                & $67.4$ \scriptsize $\pm 3.8$ & $68.8$ \scriptsize $\pm 4.0$ \\ & \VLMoai                                      & \cellcolor[HTML]{90EE90}$\textbf{75.8}$ \scriptsize $\pm 4.7$  & $60.6$ \scriptsize $\pm 6.1$                & $82.4$ \scriptsize $\pm 5.6$                                               & 72.2 \scriptsize $\pm 4.4$  & \cellcolor[HTML]{CCCCFF}\textbf{84.4} \scriptsize $\pm 4.2$                & \cellcolor[HTML]{FFCCCC}\textbf{78.2} \scriptsize $\pm 6.6$ & $81.4$ \scriptsize $\pm 4.7$ \\ & DeepSeek & $60.7$ \scriptsize $\pm 6.9$                                                        & $57.3$ \scriptsize $\pm 4.0$                & $83.0$ \scriptsize $\pm 4.6$                                               & $61.3$ \scriptsize $\pm 6.1$  & $74.8$ \scriptsize $\pm 5.6$                & $58.4$ \scriptsize $\pm 1.4$ & $69.8$ \scriptsize $\pm 2.3$ \\ \hline
\multirow{3}{*}{\normalsize $\langle \sigma_5,P_3 \rangle$} & \VLMos                                       & ---                                                                                                & $78.7$ \scriptsize $\pm 3.6$                & $89.7$ \scriptsize $\pm 0.8$                                               & $84.1$ \scriptsize $\pm 1.4$  & $90.4$ \scriptsize $\pm 1.6$                & $82.3$ \scriptsize $\pm 2.0$ & $87.7$ \scriptsize $\pm 2.1$ \\ & \VLMoai                                      & \cellcolor[HTML]{90EE90}$\textbf{87.7}$ \scriptsize $\pm 1.8$  & $85.4$ \scriptsize $\pm 1.8$                & $95.3$ \scriptsize $\pm 1.6$                                               & \cellcolor[HTML]{FFCCCC}\textbf{90.6} \scriptsize $\pm 1.5$  & \cellcolor[HTML]{CCCCFF}\textbf{95.8} \scriptsize $\pm 0.9$                & $85.5$ \scriptsize $\pm 6.1$ & $91.3$ \scriptsize $\pm 1.1$ \\ & DeepSeek  & $85.8$ \scriptsize $\pm 1.6$                                                         & $73.2$ \scriptsize $\pm 2.1$                & $91.9$ \scriptsize $\pm 0.9$                                               & $79.2$ \scriptsize $\pm 2.7$  & $89.5$ \scriptsize $\pm 1.3$                & $77.6$ \scriptsize $\pm 1.0$ & $88.6$ \scriptsize $\pm 1.4$ \\ \hline
\end{tabular}
\label{tab:resultsSentVLMs}
\end{adjustwidth}
\end{table*}

\subsection*{Direct sentiment classification using {\modelTaskOneCurto}s}
\label{Sec:exp_direct_classification}

As shown in the third column of Table~\ref{tab:resultsSentVLMs} ({\TaskOne}), the lowest $F$-scores for     {\VLMoai} and  DeepSeek occur under more challenging conditions, characterized by low annotator agreement, while higher scores are observed as annotator consensus increases. The same trend is observed with respect to the number of target classes; lower scores are obtained under harsher conditions, i.e., as the number of classes increases.    {\VLMoai} achieves higher F-scores than DeepSeek, with relative gains\footnote{For example, the relative $F$-score gain of {\VLMoai} regarding DeepSeek (DS) is calculated as (OAI - DS)/DS.} of 2\% to 67\%. 

Statistical significance was assessed using two-sided paired $t$-tests on macro-$F_1$ scores obtained through 5-fold cross validation, where both models were evaluated on identical class-balanced test partitions in each split. To control the family-wise error rate at $\alpha=0.05$ across the four configuration comparisons ($\langle \sigma_l, P_C \rangle$), Holm--Bonferroni step-down correction was applied to the raw $p$-values. The improvements are statistically significant in three configurations: both $\sigma_3$ settings ($p_{\text{adj}} < 0.001$) and $\langle \sigma_5, P_5 \rangle$ ($p_{\text{adj}} = 0.009$). In contrast, the $\langle \sigma_5, P_3 \rangle$ configuration, which exhibited the smallest absolute difference, did not reach statistical significance ($p_{\text{adj}} = 0.206$).

\begin{figure}[!htb]
    \centering
    \begin{tikzpicture}[scale=0.90, every node/.style={scale=0.90}]

\draw(0.0,3.0) node[] (img1) {\includegraphics[height=3cm,width=4cm]{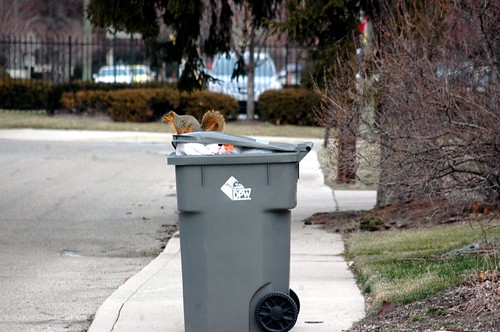}};

\node[draw, fill=white, text width=4.5cm, inner sep=2pt, align=left] (box1) at (0,-0.5) {%
  \begin{tabular}{m{3.5cm}@{}}    
   \begin{tabular}{@{}l@{}}      
      \footnotesize \textbf{Problem setup}: P5\\  [-0.2em]
      \footnotesize \textbf{Dominance threshold}: $\sigma = 3$\\ [-0.2em]
      \footnotesize \textbf{Labeled as}: Slightly Negative
    \end{tabular}
  \end{tabular}
};

\node[draw, fill=white, text width=4.5cm, inner sep=2pt, align=left, below=2pt of box1] (box2) {%
  \begin{tabular}{@{}m{0.1cm}|m{3.4cm}@{}}
    \vspace{5pt} {\rotatebox{90}{\normalsize \textbf{Task 1}}} &
    \begin{tabular}{@{}p{4.0cm}@{}}   
      \footnotesize \underline{\textbf{\VLMos}}: \fontsize{6.5pt}{7pt}\selectfont \justifying 
... Based on this image alone, it is difficult to determine the sentiment conveyed. However, assuming that the image was taken in a residential area, it is likely that the trash can is full and needs to be emptied. This could be seen as a \textbf{positive} sentiment, as it indicates that the area is well-maintained and that the residents are responsible for their waste disposal. On the other hand, if the trash can has been there for a long time and has not been emptied, it could be seen as a \textbf{negative} sentiment, indicating neglect ...
   \\ \vspace{3pt}

      \noindent \footnotesize \underline{\textbf{\VLMoai}}: Neutral\\ \vspace{3pt}
      
      \noindent \footnotesize \underline{\textbf{\VLMds}}: Neutral
    \end{tabular}
  \end{tabular}
};

\matrix(m) [matrix of math nodes,  nodes in empty cells, nodes={draw,minimum size=0.6cm, outer sep=0pt, text height=1.5ex, text depth=.15ex}] at (0.0,+1.0)
                {|[draw,fill=green!70]| \large \textbf{0} &  |[draw,fill=green!30]|  \large \textbf{0} & |[draw,fill=yellow!70]| \large \textbf{1} & |[draw,fill=red!30]| \large \textbf{3} & |[draw,fill=red!70]| \large \textbf{1}\\};

\draw(+0.0,+0.53) node[] (text0) {\scriptsize Evaluator's votes};


\end{tikzpicture}\hspace{2pt}
\begin{tikzpicture}[scale=0.90, every node/.style={scale=0.90}]

\draw(0.0,3.0) node[] (img1) {\includegraphics[height=3cm,width=4cm]{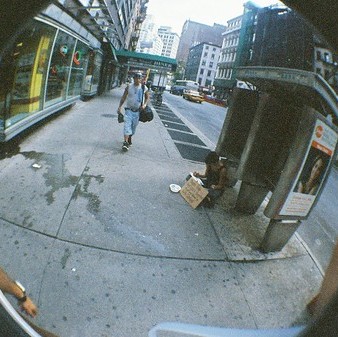}};

\node[draw, fill=white, text width=4.5cm, inner sep=2pt, align=left] (box1) at (0,-0.5) {%
  \begin{tabular}{m{3.5cm}@{}}    
   \begin{tabular}{@{}l@{}}      
      \footnotesize \textbf{Problem setup}: P3\\ [-0.2em]
      \footnotesize \textbf{Dominance threshold}: $\sigma = 3$\\  [-0.2em]
      \footnotesize \textbf{Labeled as}: Negative
    \end{tabular}
  \end{tabular}
};

\node[draw, fill=white, text width=4.5cm, inner sep=2pt, align=left, below=2pt of box1] (box2) {%
  \begin{tabular}{@{}m{0.1cm}|m{3.4cm}@{}}
    \vspace{5pt} {\rotatebox{90}{\normalsize \textbf{Task 1}}} &
    \begin{tabular}{@{}p{4.0cm}@{}}   
      \footnotesize \underline{\textbf{\VLMos}}: \fontsize{6.5pt}{7pt}\selectfont \justifying 
\textbf{Positive}: The image shows a group of people walking on the sidewalk, which suggests that they are out and about, enjoying their day or going about their daily business. The bright sunlight and clear blue sky add to the positive sentiment of the image. \textbf{Neutral}: The image is a reflection in a round mirror, making it difficult to determine the sentiment of the scene beyond the mirror ... \textbf{Negative}: It is not possible to determine the sentiment of the image due to the reflection in the round mirror. The reflection obscures the view ...
   \\ \vspace{3pt}

      \noindent \footnotesize \underline{\textbf{\VLMoai}}: Negative\\ \vspace{3pt}
      
      \noindent \footnotesize \underline{\textbf{\VLMds}}: Neutral
    \end{tabular}
  \end{tabular}
};

\matrix(m) [matrix of math nodes,  nodes in empty cells, nodes={draw,minimum size=0.6cm, outer sep=0pt, text height=1.5ex, text depth=.15ex}] at (0.0,+1.0)
{|[draw,fill=green!70]| \large \textbf{1} & |[draw,fill=green!30]| \large \textbf{0} & |[draw,fill=yellow!70]| \large \textbf{1} &  |[draw,fill=red!30]| \large \textbf{2} & |[draw,fill=red!70]| \large \textbf{1}\\};

\draw(+0.0,+0.53) node[] (text0) {\scriptsize Evaluator's votes};


\end{tikzpicture} 
    \caption{Evaluator's votes and ground-truth labels for selected images from the PerceptSent dataset, along with classification outcomes from MLLMs prompted for direct sentiment prediction (\TaskOne) under distinct problem setups and dominance thresholds. This figure is original to this work; images used within the figure are from the PerceptSent dataset~\cite{Cesar}, licensed under CC BY 4.0. Faces were blurred.}
    \label{fig:outcome-Task1}
\end{figure}

In contrast, {\VLMos} fails to produce valid classification outputs for {\TaskOne}. As illustrated in Fig.~\ref{fig:outcome-Task1}, even after extensive prompt engineering --- including explicit instructions such as ``\textit{Select a single sentiment class for this image from the $\{\mathcal{L}\}$ list}", where  $\{\mathcal{L}\}$ refers to the list of sentiment polarities for a specific problem setup, or ``\textit{Do not describe this image; just choose one sentiment polarity from the $\{\mathcal{L}\}$ list}" --- the model consistently generates explanatory responses that attempt to justify the set $\{\mathcal{L}\}$ of categories, rather than selecting a single label; or it provides  ambiguous response combining multiple sentiment cues (e.g., ``\textit{Positive, but with some neutral elements}"), making it difficult to unambiguously assign a single label and complicating automatic parsing. This behavior contrasts with the other MLLMs, which produce discrete and task-aligned outputs, as also shown in Fig.~\ref{fig:outcome-Task1}.

Therefore, we can answer research question \textbf{(Q1)} as follows: \VLMoai~is able to directly classify sentiments from raw images and consistently outperforms {\VLMds}, whereas {\VLMos} fails to produce valid or reliable outputs for this task.

\subsection*{Pre-trained LLMs for classifying MLLM image descriptions}~\label{Sec:exp_pre_trained}
\begin{figure}[!htb]
\centering
\includegraphics[scale=0.47]{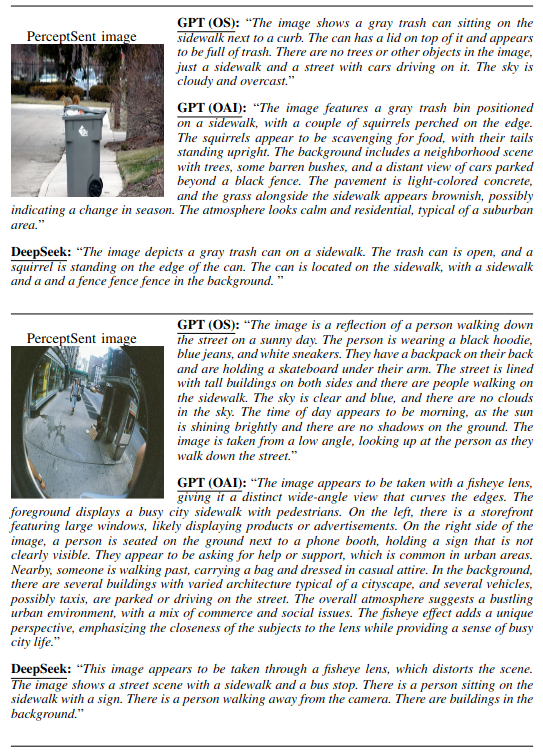}
\caption{Visual Reasoning for PerceptSent images - MLLMs descriptions (refer to Fig.~\ref{fig:outcome-Task1} for evaluator votes and target sentiment labels). This figure is original to this work; images used within the figure are from the PerceptSent dataset~\cite{Cesar}, licensed under CC BY 4.0. Faces were blurred.}
\label{fig:MLLMsDescriptions}
\end{figure}

High-quality MLLM-generated scene descriptions are fundamental to the success of \TaskTwoA, where the 
description-mediated reasoning proceeds in two stages: image description followed by a text sentiment classification. Fig.~\ref{fig:MLLMsDescriptions} illustrates two representative examples. At the top, all three models correctly identify the central subject: a gray trash can on the sidewalk. However, {\VLMos} fails to capture several salient visual elements, most notably the presence of a squirrel perched on the can, and inaccurately asserts the presence of ``cars driving on the street,''  while also stating that no other objects are visible, thereby overlooking parked vehicles and surrounding vegetation. {\VLMoai} produces a more detailed account of the scene, identifying the squirrel (although it incorrectly refers to two animals), describing it as likely scavenging for food in the trash can, an action commonly associated with negative sentiment, and including broader contextual cues such as barren bushes, seasonal grass tones, and the calm suburban environment. {\VLMds} also detects the squirrel and its position on the edge of the can. Still, its output suffers from significant redundancy, including repeated phrases such as ``a sidewalk and a and a fence fence fence," a recurring issue that reduces clarity and readability. In the bottom image, both {\VLMos} and {\VLMds} produce short, less context-aware descriptions, failing to capture crucial semantic elements. In contrast, {\VLMoai} is the only model to correctly identify a person sitting on the floor asking for help, an observation also made by some human annotators, whose descriptions  have included terms such as beggar and poverty, both strongly associated with the image's negative sentiment label.

Therefore, as reported in Table~\ref{tab:resultsSentVLMs} (\TaskTwoA),   {\VLMoai}  achieves the highest $F$-scores across  all $\langle \sigma_l,P_C \rangle$ and LLM architectures. Although {\VLMds} generally scores lower in {\TaskTwoA} compared with {\VLMos} and {\VLMoai}, with   BART for $\sigma_3$, or MBERT in all configurations, it shows consistent improvements from {\TaskOne} to {\TaskTwoA}.

Overall, MLLM+LLM performance benefits more from higher annotator consensus than from reduced class granularity. Whereas decreasing the number of sentiment categories from $P_5$ to $P_3$ leads to improvements, increasing the consensus level (from $\sigma_3$ to $\sigma_5$) mostly results in more substantial gains. The {\VLMoai}+MBERT model in {\TaskTwoA}, under $P_5$ for example, improves from an average $F$-score $47.9$ at $\sigma_3$ to $72.2$ at $\sigma_5$, a relative gain of $50.7\%$. A similar pattern is observed for {\VLMds} and {\VLMos}, whose MBERT-based pipelines improve from an average $F$-score of  $38.2$ to $61.3$, and $43.4$ to $65.4$, respectively. 

Based on the previous results, the answer to research question \textbf {(Q2)} depends on the LLM: using MBERT to classify textual {\imgDescrip}s leads to better results except in one case; BART and LLAMA show mixed results. Moreover, adding LLMs enables classification for {\VLMos}, which could not perform sentiment analysis on raw images.
These results suggest that the image-to-text transformation may help reorganize visual information into a representation that better aligns with the reasoning capabilities of text-based LLMs.

\subsection*{Fine-tuning LLMs for classifying MLLM image descriptions}
\label{Sec:exp_fine_tuning}

Fine-tuning the LLMs ({\TaskTwoB} in Table~\ref{tab:resultsSentVLMs}) yields the highest overall $F$-scores. Relative gains over {\TaskTwoA} range from [11.59\%, 50.87\%] for BART, [5.74\%, 32.20\%] for MBERT, and [4.09\%, 102.87\%] for LLAMA, across all MLLMs, evaluators' agreement thresholds, and number of sentiment classes.

These results suggest that parameter adaptation broadly enhances sentiment reasoning from multimodal input, independent of the task's complexity, and allows us to answer the research question \textbf{(Q3)}: fine-tuning LLMs for sentiment classification positively impacts the overall results.

\begin{figure*}[htb!]
    \centering
    \begin{tikzpicture}[scale=0.88, every node/.style={scale=0.88}]
                \scriptsize
                \draw(0.0,3.0) node[] (img1) {\includegraphics[height=3cm,width=3cm]{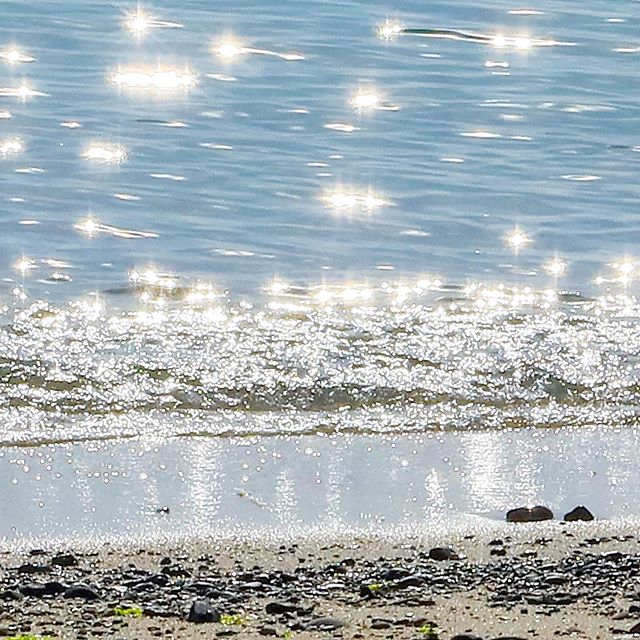}};

                \node[draw, fill=white, text width=7.7cm, inner sep=2pt, align=left] (box1) at ([xshift=4.6cm, yshift=1.0cm]img1.east) {
                \begin{tabular}{m{3.5cm}@{}}
                \begin{tabular}{@{}l@{}}
                \footnotesize \textbf{Problem setup}: P5\\
                \footnotesize \textbf{Dominance threshold}: $\sigma_5$\\
                \footnotesize \textbf{Labeled as}: Positive\\
                \footnotesize \textbf{Human perceptions (\# of votes)}:\\ \footnotesize colors (1),  \footnotesize leisure/fun/rest (1), \footnotesize sky (1), 
                \footnotesize nature (4), \\
                \footnotesize pleasant environment (3).
                \end{tabular}
                \end{tabular}
                };

                \node[draw, fill=white, text width=7.8cm, inner sep=1pt, align=left, below=2pt of box1] (box2) {%
                \setlength{\tabcolsep}{2pt}
                \renewcommand{\arraystretch}{1.3}  
                \begin{tabular}{clccc}
                &  & {\VLMos} & {\VLMoai} & {\VLMds} \\ \cline{3-5}
                \multirow{2}{*}{\underline{BART}}  & \footnotesize \textbf{Task 2$_a$:} &  Positive & Positive & Slightly Pos. \\
                & \footnotesize \textbf{Task 2$_b$:} & Positive & Positive & Positive \\ \cline{2-5}
                \multirow{2}{*}{\underline{MBERT}}  & \footnotesize \textbf{Task 2$_a$:} &  Slightly Pos. & Positive & Positive \\
                & \footnotesize \textbf{Task 2$_b$:} & Positive & Positive & Positive \\ \cline{2-5}
                \multirow{2}{*}{\underline{LLAMA}}  & \footnotesize \textbf{Task 2$_a$:} & Positive & Positive & Positive \\
                & \footnotesize \textbf{Task 2$_b$:} &  Positive & Positive & Positive \\
                \end{tabular}
                };
                \matrix(m) [matrix of math nodes,  nodes in empty cells, nodes={draw,minimum size=0.6cm, outer sep=0pt, text height=1.5ex, text depth=.15ex, font=\large}] at (0.0,+1.0)
                {|[draw,fill=green!70]|  \textbf{5} &  |[draw,fill=green!30]|   \textbf{0} & |[draw,fill=yellow!70]| \textbf{0} & |[draw,fill=red!30]|  \textbf{0} & |[draw,fill=red!70]| \textbf{0}\\};

                \draw(+0.0,+0.53) node[] (text0) {\scriptsize Evaluator's votes};

                \draw node[below=2pt of box2] (caption) {\normalsize (a)};
                \end{tikzpicture}
                
 \vspace{3pt}

\begin{tikzpicture}[scale=0.88, every node/.style={scale=0.88}]
                \scriptsize
                \draw(0.0,3.0) node[] (img1) {\includegraphics[height=3cm,width=3cm]{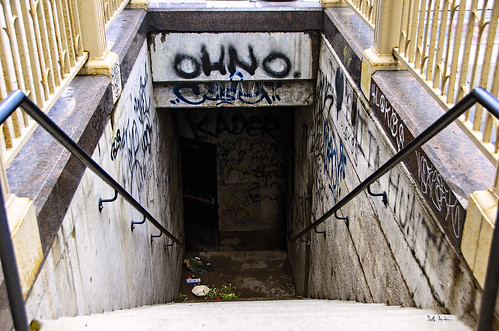}};

                \node[draw, fill=white, text width=7.7cm, inner sep=2pt, align=left] (box1) at ([xshift=4.6cm, yshift=1.0cm]img1.east) {
                \begin{tabular}{m{3.5cm}@{}}
                \begin{tabular}{@{}l@{}}
                \footnotesize \textbf{Problem setup}: P3\\
                \footnotesize \textbf{Dominance threshold}: $\sigma_5$\\
                \footnotesize \textbf{Labeled as}: Negative\\
                \footnotesize \textbf{Human perceptions (\# of votes)}:\\ 
                \footnotesize lack of maintenance (3),
                \footnotesize trash (1)
                \footnotesize graffiti (5), \footnotesize violence (1),\\
                \footnotesize negative text message (1) 
                \footnotesize debris/destruction (1)
                
                \end{tabular}
                \end{tabular}
                };

                \node[draw, fill=white, text width=7.8cm, inner sep=1pt, align=left, below=2pt of box1] (box2) {%
                \setlength{\tabcolsep}{2pt}
                \renewcommand{\arraystretch}{1.3}  
                \begin{tabular}{clccc}
                &  & {\VLMos} & {\VLMoai} & {\VLMds} \\ \cline{3-5}
                \multirow{2}{*}{\underline{BART}}  & \footnotesize \textbf{Task 2$_a$:} &  Negative & Negative & Negative \\
                & \footnotesize \textbf{Task 2$_b$:} &  Negative & Negative & Negative \\ \cline{2-5}
                \multirow{2}{*}{\underline{MBERT}}  & \footnotesize \textbf{Task 2$_a$:} &  Negative & Negative & Negative \\
                & \footnotesize \textbf{Task 2$_b$:} &  Negative & Negative & Negative \\ \cline{2-5}
                \multirow{2}{*}{\underline{LLAMA}}  & \footnotesize \textbf{Task 2$_a$:} &  Negative & Negative & Negative \\
                & \footnotesize \textbf{Task 2$_b$:} & Negative & Negative & Negative \\
                \end{tabular}
                };
                \matrix(m) [matrix of math nodes,  nodes in empty cells, nodes={draw,minimum size=0.6cm, outer sep=0pt, text height=1.5ex, text depth=.15ex, font=\large}] at (0.0,+1.0)
                {|[draw,fill=green!70]| \large \textbf{0} &  |[draw,fill=green!30]|  \large \textbf{0} & |[draw,fill=yellow!70]| \large \textbf{0} & |[draw,fill=red!30]| \large \textbf{1} & |[draw,fill=red!70]| \large \textbf{4}\\};

                \draw(+0.0,+0.53) node[] (text0) {\scriptsize Evaluator's votes};

                \draw node[below=2pt of box2] (caption) {\normalsize (b)};
                \end{tikzpicture}

 \vspace{3pt}

\begin{tikzpicture}[scale=0.88, every node/.style={scale=0.88}]
                \scriptsize
                \draw(0.0,3.0) node[] (img1) {\includegraphics[height=3cm,width=3cm]{174EgdAhI2ZAMmKXmv42SgJDN7Wm9rc.jpg}};

                \node[draw, fill=white, text width=7.8cm, inner sep=2pt, align=left] (box1) at ([xshift=4.7cm, yshift=1.0cm]img1.east) {
                \begin{tabular}{m{3.5cm}@{}}
                \begin{tabular}{@{}l@{}}
                \footnotesize \textbf{Problem setup}: P5\\
                \footnotesize \textbf{Dominance threshold}: $\sigma_3$\\
                \footnotesize \textbf{Labeled as}: Slightly Negative\\
                \footnotesize \textbf{Human perceptions (\# of votes)}:\\ 
                \footnotesize \underline{trash} (4), 
                \footnotesize pollution (1),
                \footnotesize debris/destruction (1), \\
                \footnotesize lack of maintenance (1), 
                \footnotesize animals (1), 
                \footnotesize everyday image (1)
                \footnotesize
                \end{tabular}
                \end{tabular}
                };

                \node[draw, fill=white, text width=7.9cm, inner sep=1pt, align=left, below=2pt of box1] (box2) {%
                \setlength{\tabcolsep}{2pt}
                \renewcommand{\arraystretch}{1.3}  
                \begin{tabular}{clccc}
                &  & {\VLMos} & {\VLMoai} & {\VLMds} \\ \cline{3-5}
                \multirow{2}{*}{\underline{BART}}  & \footnotesize \textbf{Task 2$_a$:} & Slightly Neg. & Slightly Neg. & Slightly Neg. \\
                & \footnotesize \textbf{Task 2$_b$:} &  Negative & Neutral & Slightly Neg. \\ \cline{2-5}
                \multirow{2}{*}{\underline{MBERT}}  & \footnotesize \textbf{Task 2$_a$:} &  Slightly Neg. & Slightly Neg. & Negative \\
                & \footnotesize \textbf{Task 2$_b$:} &  Slightly Neg. & Slightly Neg. & Slightly Neg. \\ \cline{2-5}
                \multirow{2}{*}{\underline{LLAMA}}  & \footnotesize \textbf{Task 2$_a$:} &  Positive & Neutral & Slightly Neg. \\
                & \footnotesize \textbf{Task 2$_b$:} &  Slightly Neg. & Neutral & Slightly Neg. \\
                \end{tabular}
                };
                \matrix(m) [matrix of math nodes,  nodes in empty cells, nodes={draw,minimum size=0.6cm, outer sep=0pt, text height=1.5ex, text depth=.15ex, font=\large}] at (0.0,+1.0)
                {|[draw,fill=green!70]| \large \textbf{0} &  |[draw,fill=green!30]|  \large \textbf{0} & |[draw,fill=yellow!70]| \large \textbf{1} & |[draw,fill=red!30]| \large \textbf{3} & |[draw,fill=red!70]| \large \textbf{1}\\};

                \draw(+0.0,+0.53) node[] (text0) {\scriptsize Evaluator's votes};

                \draw node[below=2pt of box2] (caption) {\normalsize (c)};
                \end{tikzpicture}
                
 \vspace{3pt}

\begin{tikzpicture}[scale=0.88, every node/.style={scale=0.88}]
                \scriptsize
                \draw(0.0,3.0) node[] (img1) {\includegraphics[height=3cm,width=3cm]{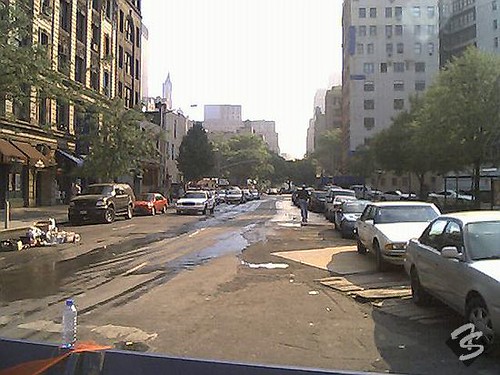}};

                \node[draw, fill=white, text width=7.8cm, inner sep=2pt, align=left] (box1) at ([xshift=4.7cm, yshift=1.0cm]img1.east) {
                \begin{tabular}{m{3.5cm}@{}}
                \begin{tabular}{@{}l@{}}
                \footnotesize \textbf{Problem setup}: P3\\
                \footnotesize \textbf{Dominance threshold}: $\sigma_3$\\
                \footnotesize \textbf{Labeled as}: Negative\\
                \footnotesize \textbf{Human perceptions (\# of votes)}:\\ 
                \footnotesize trash (2), 
                \footnotesize everyday image (1),
                \footnotesize accident (1), \\
                \footnotesize lack of colors (1), 
                \footnotesize pollution (1),
                \footnotesize lack of maintenance (1) 
                \footnotesize
                \end{tabular}
                \end{tabular}
                };

                \node[draw, fill=white, text width=7.9cm, inner sep=1pt, align=left, below=2pt of box1] (box2) {%
                \setlength{\tabcolsep}{4pt}
                \renewcommand{\arraystretch}{1.3}  
                \begin{tabular}{clccc}
                &  & {\VLMos} & {\VLMoai} & {\VLMds} \\ \cline{3-5}
                \multirow{2}{*}{\underline{BART}}  & \footnotesize \textbf{Task 2$_a$:} &  Positive & Neutral & Positive \\
                & \footnotesize \textbf{Task 2$_b$:} &  Neutral & Neutral & Positive \\ \cline{2-5}
                \multirow{2}{*}{\underline{MBERT}}  & \footnotesize \textbf{Task 2$_a$:} &  Neutral & Neutral & Neutral \\
                & \footnotesize \textbf{Task 2$_b$:} & Neutral & Neutral & Neutral \\ \cline{2-5}
                \multirow{2}{*}{\underline{LLAMA}}  & \footnotesize \textbf{Task 2$_a$:} & Negative & Negative & Negative \\
                & \footnotesize \textbf{Task 2$_b$:} &  Neutral & Neutral & Neutral \\
                \end{tabular}
                };
                \matrix(m) [matrix of math nodes,  nodes in empty cells, nodes={draw,minimum size=0.6cm, outer sep=0pt, text height=1.5ex, text depth=.15ex, font=\large}] at (0.0,+1.0)
                {|[draw,fill=green!70]| \large \textbf{0} &  |[draw,fill=green!30]|  \large \textbf{0} & |[draw,fill=yellow!70]| \large \textbf{2} & |[draw,fill=red!30]| \large \textbf{1} & |[draw,fill=red!70]| \large \textbf{2}\\};

                \draw(+0.0,+0.53) node[] (text0) {\scriptsize Evaluator's votes};

                \draw node[below=2pt of box2] (caption) {\normalsize (d)};
                \end{tikzpicture}
    \caption{Qualitative comparison of sentiment predictions on
PerceptSent examples using MLLM and LLM combinations for
\textbf{Task~2\textsubscript{a}} and \textbf{Task~2\textsubscript{b}}, under different setups and evaluator agreement. This figure is original to this work; images used within the figure are from the PerceptSent dataset~\cite{Cesar}, licensed under CC BY 4.0.}
    \label{fig:tasks}
\end{figure*}

Fig.~\ref{fig:tasks} showcases four examples from the PerceptSent dataset alongside sentiment predictions from all considered models in both {\TaskTwoA} and {\TaskTwoB}. 
 Figs.~\ref{fig:tasks}(a,b) present image samples with strong consensus ($\sigma_5$) among human evaluators. In Fig.~\ref{fig:tasks}(a), the image has received unanimous positive sentiment annotations from all five annotators. The associated human perceptions are strongly positive, with descriptions such as nature (mentioned four times) and pleasant environment (three times), among others. After fine-tuning, all MLLM+LLM combinations correctly predict the positive sentiment based on generated descriptions referencing visual elements such as a body of water, a beach, and sparkling highlights. 
 
Fig.~\ref{fig:tasks}(b) similarly shows high agreement, with  human evaluators associating the negative sentiment with the presence of graffiti (noted by all annotators), lack of maintenance (three mentions), destruction (one mention), among other negative perceptions. 
 The  MLLMs {\captioning} results often capture these perceptions, highlighting features such as an underground tunnel, a graffiti-covered entrance, and a gritty atmosphere, with {\VLMoai} explicitly mentioning a sense of decay.  Therefore, all MLLM+LLM combinations (pre-trained and fine-tuned) accurately identify the target negative sentiment. Interestingly, both {\VLMoai} and {\VLMds} can even recognize and transcribe the ``OHNO" graffiti above the entrance. As shown in Table~\ref{tab:resultsSentVLMs}, these two cases exemplify the best observed performance, where high evaluator agreement ($\sigma_5$) correlates with elevated $F$-scores. 
 
 In contrast, under lower agreement ($\sigma_3$), some issues occur.   Although in Fig.~\ref{fig:tasks}(c) most of the pre-trained models correctly classify slightly negative sentiment, some BART predictions shift toward wrong sentiment after fine-tuning. This change may result from the influence of {\captioning}s generated by {\VLMoai}, such as ``The atmosphere looks calm and residential, typical of a suburban area," which potentially moderates the models' sentiment evaluations. Nevertheless, fine-tuning improvements occur, particularly for \VLMos+LLAMA and \VLMds+MBERT.

In the case illustrated by Fig.\ref{fig:tasks}(d),   none of the MLLM+LLM combinations succeed in predicting the correct negative sentiment label after fine-tuning (\TaskTwoB). This outcome is likely driven by limitations in the scene descriptions generated by the MLLMs. Specifically, none of the models could identify salient visual features indicative of a negative context --- such as the pile of trash on the left or the overall untidy condition of the street, as indicated in the human evaluator perceptions (marked as trash (two times), pollution, and lack of maintenance). Instead, the generated {\imgDescrip}s emphasize neutral or positive aspects, including references to a city street with parked cars, a calm atmosphere, and the presence of a few pedestrians. Such framing has likely influenced the LLMs to interpret the scene as neutral. Notably, LLAMA in {\TaskTwoA} predicts the correct negative sentiment; however, with low $F$-scores ($[30.3\%,46.6\%]$) for $\langle \sigma_3, P_3 \rangle$, this does not indicate consistent performance. After fine-tuning, the prediction shifted to neutral, confirming that the generated descriptions failed to capture the negative elements identified by human evaluators.

\subsection*{Final remarks on MLLMs: performance sensitivity and behavioral analysis} \label{sec:FinalRemarks}
As shown in Table~\ref{tab:resultsSentVLMs}, all evaluated models exhibit sensitivity to both annotator agreement ($\sigma_3$ or $\sigma_5$) and class granularity ($P_3$ or $P_5$). More specifically, performance consistently decreases in settings associated with lower annotator consensus ($\sigma_3$), where the sentiment interpretation of the images becomes more ambiguous and subjective. A similar degradation is observed in fine-grained classification scenarios ($P_5$), in which the models must discriminate among a larger number of sentiment categories with subtler semantic differences. These results suggest that both label uncertainty and increased class complexity substantially affect the difficulty of the task.

Beyond this overall tendency, our experiments also indicate that differences in model architecture and response behavior significantly influence performance. For example, GPT (OS) frequently produces explanatory, descriptive, or multi-label outputs even when the prompts explicitly request a single constrained classification label. This behavior reduces its effectiveness in classification-oriented settings because the generated responses become less consistent with the expected output format. In contrast, GPT (OAI) and DeepSeek tend to produce more structured and task-aligned outputs, adhering more closely to the required classification protocol.

Our qualitative analyses further indicate that higher-performing models are generally more capable of identifying and verbalizing contextual and scene-level semantic information in their generated descriptions. This capability appears particularly important in image sentiment analysis, where emotional interpretation often depends not only on isolated visual objects, but also on subtle contextual interactions, social situations, and affective cues distributed throughout the scene.

Finally, to assess whether the observed trends generalize across additional recent MLLM families, S3 Appendix presents supplementary experiments using Gemini, Phi-4, and Gemma-4-E4B under the same evaluation protocol. The results reinforce the central findings of this work, particularly the effectiveness and robustness of description-mediated reasoning combined with fine-tuned MBERT.


\subsection*{{\propAppCurto} \textit{vs} language and image-based baselines}
\label{Sec:exp_vs_other_classifiers}

To focus the discussion, we refer to the \textbf{GPT (OAI) + MBERT} configuration --- the top performer in Table~\ref{tab:resultsSentVLMs}, Task~2\textsubscript{b} --- as \textbf{\propAppCurto}, here and in the following sections, which we treat as the reference instantiation of our framework. 

As shown in Fig.~\ref{fig:plotBaselinesAndSentVLMs}, we benchmark the {\propAppCurto} performance against three baselines designed to capture sentiment through different modeling paradigms. The first is VADER \cite{hutto2014vader}, a widely used rule- and lexicon-based sentiment analysis tool originally designed for processing emotive content in social media and informal text. The second and third baselines are deep learning methods that operate directly on images: a ResNet-based CNN model, as proposed by \cite{outdoorsent2020},  and one adapted for this work,  a Swin Transformer model. Usually, Transformers leverage hierarchical attention to capture visual patterns more effectively than conventional CNNs. The analysis that follows provides insights into how architectural choices and input modality --- textual versus visual --- affect performance across varying sentiment granularities and levels of annotator agreement.

\textbf{CNN vs. Transformer Performance:} While ResNet shows marginally higher point estimates in coarse classification ($\sigma_3$, Figs. \ref{fig:plotBaselinesAndSentVLMs}a-b), the overlapping confidence intervals with Swin Transformer suggest this difference may not be statistically significant. However, Swin Transformer shows clearer advantages in fine-grained scenarios ($\sigma_5$, Figs. \ref{fig:plotBaselinesAndSentVLMs}c-d), where its attention mechanisms better capture image details and, consequently, the consensus sentiment cues.  
\begin{figure}[!htb]
\centering
\begin{tikzpicture}[scale=0.85]
\begin{axis}[
    height=3.5cm,
    width=10.7cm,
    ybar=0pt,
    bar width=16pt,
    ymin=0,
    ymax=100,
    enlarge x limits=0.15,
    ylabel={\large F-score},
    xlabel={\large (a) Sentiment Polarity Setup: $\langle \sigma_3, P_5 \rangle$},
    xtick={0,1,2,3},
    xticklabel style={text width=2.5cm, align=center},
    xticklabels={
    \normalsize ,
    \normalsize ResNet\\CNN,
    \normalsize Swin\\Transformer,
    \normalsize {\propAppCurto}
    },
    ylabel style={yshift=-6pt},
    ymajorgrids=true,
    ytick={0,20,40,60,80,100},
    bar shift=0pt 
]

\addplot+[
    style={black, fill=gray!20, postaction={pattern=north east lines}}
] plot [error bars/.cd, y dir=both, y explicit] coordinates {(0,0.0) +- (0,0.0)};

\addplot+[
    style={black, fill=gray!40, postaction={pattern=north west lines}}
] plot [error bars/.cd, y dir=both, y explicit] coordinates {(1,45.0) +- (0,3.4)};

\addplot+[
    style={black, fill=gray!20, postaction={pattern=crosshatch}}
] plot [error bars/.cd, y dir=both, y explicit] coordinates {(2,41.01) +- (0,2.7)};

\addplot+[
    style={black, fill=blue!30, postaction={pattern=north east lines}}
] plot [error bars/.cd, y dir=both, y explicit] coordinates {(3,58.4) +- (0,4.0)};

\end{axis}
\node[anchor=south west] (img) at (0.32,+1.91) {\small 95\% confidence interval}; 
\draw[black] (0.3,+2.30) -- (0.3,+2.00);
\draw[black] (0.2,+2.30) -- (0.4,+2.30);
\draw[black] (0.2,+2.00) -- (0.4,+2.00);
\end{tikzpicture}
\begin{tikzpicture}[scale=0.85]
\begin{axis}[
    height=3.5cm,
    width=10.7cm,
    ybar=0pt,
    bar width=16pt,
    ymin=0,
    ymax=100,
    enlarge x limits=0.15,
    ylabel={\large F-score},
    xlabel={\large (b) Sentiment Polarity Setup: $\langle \sigma_3, P_3 \rangle$},
    xtick={0,1,2,3},
    xticklabel style={text width=2.5cm, align=center},
    xticklabels={
    \normalsize VADER,
    \normalsize ResNet\\CNN,
    \normalsize Swin\\Transformer,
    \normalsize {\propAppCurto}
    },
    ylabel style={yshift=-6pt},
    ymajorgrids=true,
    ytick={0,20,40,60,80,100},
    bar shift=0pt 
]

\addplot+[
    style={black, fill=gray!20, postaction={pattern=north east lines}}
] plot [error bars/.cd, y dir=both, y explicit] coordinates {(0,59.19) +- (0,1.27)};

\addplot+[
    style={black, fill=gray!40, postaction={pattern=north west lines}}
] plot [error bars/.cd, y dir=both, y explicit] coordinates {(1,61.0) +- (0,5.3)};

\addplot+[
    style={black, fill=gray!20, postaction={pattern=crosshatch}}
] plot [error bars/.cd, y dir=both, y explicit] coordinates {(2,59.51) +- (0,5.0)};

\addplot+[
    style={black, fill=blue!30, postaction={pattern=north east lines}}
] plot [error bars/.cd, y dir=both, y explicit] coordinates {(3,77.5) +- (0,0.9)};

\end{axis}
\node[anchor=south west] (img) at (0.32,+1.91) {\small 95\% confidence interval}; 
\draw[black] (0.3,+2.30) -- (0.3,+2.00);
\draw[black] (0.2,+2.30) -- (0.4,+2.30);
\draw[black] (0.2,+2.00) -- (0.4,+2.00);
\end{tikzpicture}

\vspace{5pt}
\begin{tikzpicture}[scale=0.85]
\begin{axis}[
    height=3.5cm,
    width=10.7cm,
    ybar=0pt,
    bar width=16pt,
    ymin=0,
    ymax=100,
    enlarge x limits=0.15,
    ylabel={\large F-score},
    xlabel={\large (c) Sentiment Polarity Setup: $\langle \sigma_5, P_5 \rangle$},
    xtick={0,1,2,3},
    xticklabel style={text width=2.5cm, align=center},
    xticklabels={
    \normalsize ,
    \normalsize ResNet\\CNN,
    \normalsize Swin\\Transformer,
    \normalsize {\propAppCurto}
    },
    ylabel style={yshift=-6pt},
    ymajorgrids=true,
    ytick={0,20,40,60,80,100},
    bar shift=0pt
]

\addplot+[
    style={black, fill=gray!20, postaction={pattern=north east lines}}
] plot [error bars/.cd, y dir=both, y explicit] coordinates {(0,0.0) +- (0,0.0)};

\addplot+[
    style={black, fill=gray!40, postaction={pattern=north west lines}}
] plot [error bars/.cd, y dir=both, y explicit] coordinates {(1,51.2) +- (0,6.3)};

\addplot+[
    style={black, fill=gray!20, postaction={pattern=crosshatch}}
] plot [error bars/.cd, y dir=both, y explicit] coordinates {(2,64.33) +- (0,4.8)};

\addplot+[
    style={black, fill=blue!30, postaction={pattern=north east lines}}
] plot [error bars/.cd, y dir=both, y explicit] coordinates {(3,84.4) +- (0,4.2)};

\end{axis}
\node[anchor=south west] (img) at (0.32,+1.91) {\small 95\% confidence interval}; 
\draw[black] (0.3,+2.30) -- (0.3,+2.00);
\draw[black] (0.2,+2.30) -- (0.4,+2.30);
\draw[black] (0.2,+2.00) -- (0.4,+2.00);
\end{tikzpicture}
\begin{tikzpicture}[scale=0.85]
\begin{axis}[
    height=3.5cm,
    width=10.7cm,
    ybar=0pt,
    bar width=16pt,
    ymin=0,
    ymax=100,
    enlarge x limits=0.15,
    ylabel={\large F-score},
    xlabel={\large (d) Sentiment Polarity Setup: $\langle \sigma_5, P_3 \rangle$},
    xtick={0,1,2,3},
    xticklabel style={text width=2.5cm, align=center},
    xticklabels={
    \normalsize VADER,
    \normalsize ResNet\\CNN,
    \normalsize Swin\\Transformer,
    \normalsize {\propAppCurto}
    },
    ylabel style={yshift=-6pt},
    ymajorgrids=true,
    ytick={0,20,40,60,80,100},
    bar shift=0pt 
]

\addplot+[
    style={black, fill=gray!20, postaction={pattern=north east lines}}
] plot [error bars/.cd, y dir=both, y explicit] coordinates {(0,82.21) +- (0,1.69)};

\addplot+[
    style={black, fill=gray!40, postaction={pattern=north west lines}}
] plot [error bars/.cd, y dir=both, y explicit] coordinates {(1,69.0) +- (0,4.7)};

\addplot+[
    style={black, fill=gray!20, postaction={pattern=crosshatch}}
] plot [error bars/.cd, y dir=both, y explicit] coordinates {(2,79.2) +- (0,2.7)};

\addplot+[
    style={black, fill=blue!30, postaction={pattern=north east lines}}
] plot [error bars/.cd, y dir=both, y explicit] coordinates {(3,95.8) +- (0,0.9)};

\end{axis}
\node[anchor=south west] (img) at (0.32,+1.91) {\small 95\% confidence interval}; 
\draw[black] (0.3,+2.30) -- (0.3,+2.00);
\draw[black] (0.2,+2.30) -- (0.4,+2.30);
\draw[black] (0.2,+2.00) -- (0.4,+2.00);
\end{tikzpicture}
\caption{$F$-score results for image sentiment polarity classification on the PerceptSent dataset under different setups. Each subfigure represents a distinct configuration of sentiment granularity ($\sigma_l$) and polarity class count ($P_C$). The ResNet CNN results correspond to the architecture described in~\cite{PerceptSent}.}
\label{fig:plotBaselinesAndSentVLMs}
\end{figure}

\textbf{VADER vs. Deep Learning:}  We include the rule-based VADER model in our analysis by considering the same MLLM-generated {\imgDescrip}s previously used as inputs in {\TaskTwo}. VADER computes a compound score, i.e., a normalized metric in the range $[-1,1]$ by aggregating the valence of each word in the input text. Adapting its standard configuration, we assign sentiment labels based on fixed thresholds: positive if the compound score $> 0.5$, negative if $< -0.5$, and neutral otherwise. Unlike learning-based methods, VADER cannot be systematically retrained or fine-tuned to capture the nuances of richer, domain-specific descriptions. As shown in Figs.~\ref{fig:plotBaselinesAndSentVLMs}(b,d), we evaluate VADER only on the $P_3$ problems, since its predefined rules are limited to coarse sentiment categories (negative, neutral, and positive), making it unsuitable for the more fine-grained sentiment distinctions required in the $P_5$ setting. Despite its simplicity and lack of training, VADER achieves competitive performance—comparable to ResNet and Transformer-based models for $\langle \sigma_3, P_3 \rangle$, and even slightly superior for $\langle \sigma_5, P_3 \rangle$—highlighting the surprising effectiveness of rule-based sentiment analysis when paired with high-quality textual input. 

\textbf{{\propAppCurto} performance relative to baselines:}
To make this comparison explicit, we conducted a targeted post-hoc analysis comparing {\propAppCurto} with the baselines reported in Fig.~\ref{fig:plotBaselinesAndSentVLMs}. For Swin Transformer and VADER, we used two-sided paired $t$-tests over the five cross-validation folds. For ResNet CNN, whose fold-wise outputs were unavailable, we used an approximate Welch-type comparison based on the reported mean and 95\% confidence interval. Holm--Bonferroni correction was applied across the 10 comparisons. The analysis indicates that {\propAppCurto} remains significantly higher than the Swin Transformer, ResNet CNN, and VADER baselines in every applicable configuration (adjusted $p_{\mathrm{Holm}}\leq 0.0012$). 

Unlike conventional CNN- or Transformer-based visual classifiers, which directly infer sentiment from visual representations, the proposed pipeline decomposes the task into scene interpretation followed by linguistic sentiment reasoning, enabling the model to capture contextual and socially grounded affective cues.

Rather than reflecting architectural novelty, these results suggest that the MLLM description-mediated pipeline studied (where the best configuration is referred to as {\propAppCurto}) is particularly effective in scenarios where sentiment depends on contextual or scene-level cues. The image description generated by the {\VLMoai} MLLM  preserves both visual content and affective cues, enabling more accurate sentiment reasoning. For instance, textual descriptions can externalize contextual elements (e.g., references to isolation, decay, or social setting) that align more closely with how annotators justify their sentiment judgments.

To further illustrate the interpretability of the proposed description-mediated pipeline, Table~\ref{tab:interpretability_examples} presents representative examples in which the generated descriptions, using GPT (OAI), expose semantic and affective cues that align with human sentiment perceptions (available on the PerceptSent dataset~\cite{Cesar}).

\begin{table*}[t]
\begin{adjustwidth}{-2.25in}{0in}
\scriptsize
\centering

\caption{Representative examples illustrating how the proposed description-mediated pipeline externalizes semantic and affective cues through MLLM-generated descriptions, using GPT (OAI). Correct predictions refer to the problem $\langle \sigma_5,P_5 \rangle$. The extracted cues align closely with human perceptual interpretations and help explain the downstream sentiment predictions. Images illustrated in this table are from the PerceptSent dataset~\cite{Cesar}, licensed under CC BY 4.0. Faces were blurred. 
}

\begin{tabular}{
m{1.9cm}
m{5.5cm}
m{3.0cm}
m{2.5cm}
m{3.4cm}
}

\hline

\textbf{Image} &
\textbf{MLLM description (excerpt)} &
\textbf{Semantic cues influencing sentiment} &
\textbf{Groundtruth/ Predicted} &
\textbf{Human perception cues} \\

\hline

{\centering\includegraphics[width=1.5cm]{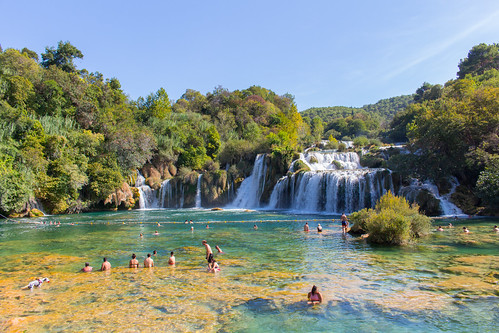}\par} &
``beautiful waterfall surrounded by lush greenery'', ``people are swimming and enjoying the coolness of the water'', ``tranquil and inviting atmosphere'', ``people engaging in leisure activities'' &
nature, leisure, relaxation, vibrant colors, tourism &
Positive/Positive &
``Pleasant environment'', ``Leisure/Fun/Rest'', ``Nature'', ``Colors'', ``Tourist Attractions'' \\

\hline

{\centering\includegraphics[width=1.5cm]{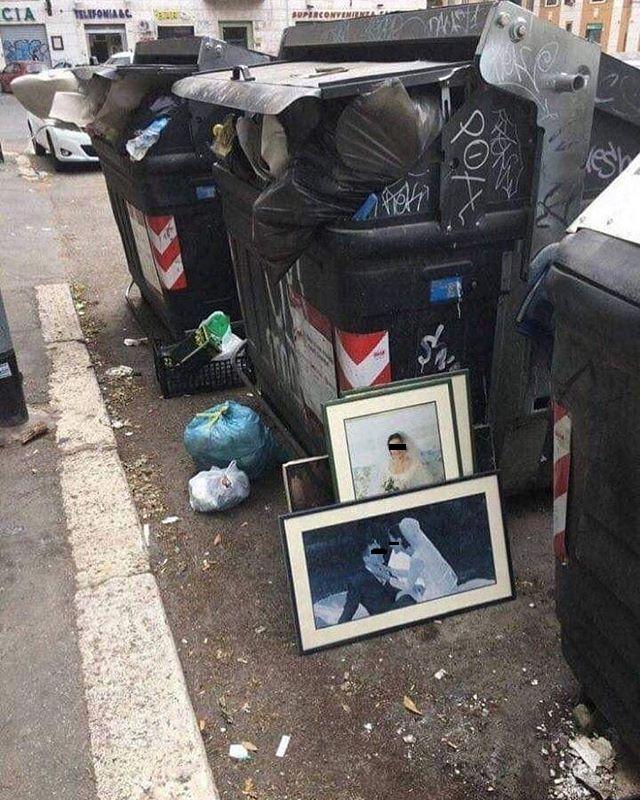}\par} &
``overflowing garbage bins'', ``the area around the bins is littered with various pieces of garbage'', ``neglected or low-maintenance environment'', ``sense of abandonment'' &
trash, neglect, abandonment, urban degradation &
Negative/Negative &
``Lack of Maintenance'', ``Trash'', ``Sadness'' \\

\hline

{\centering\includegraphics[width=1.5cm]{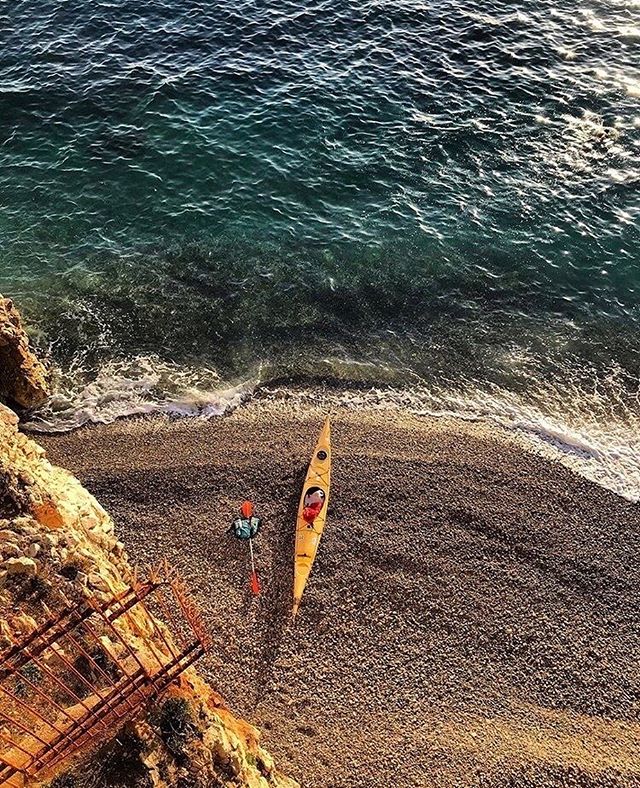}\par} &
``serene coastal scene'', ``beautiful shade of blue'', ``peaceful solitude in nature'', ``perfect for kayaking and enjoying the outdoors'' &
calmness, nature, sports, leisure, panoramic scenery &
Positive/Positive &
``Pleasant environment'', ``Nature'', ``Sports'', ``Leisure/Fun/Rest'' \\

\hline

{\centering\includegraphics[width=1.5cm]{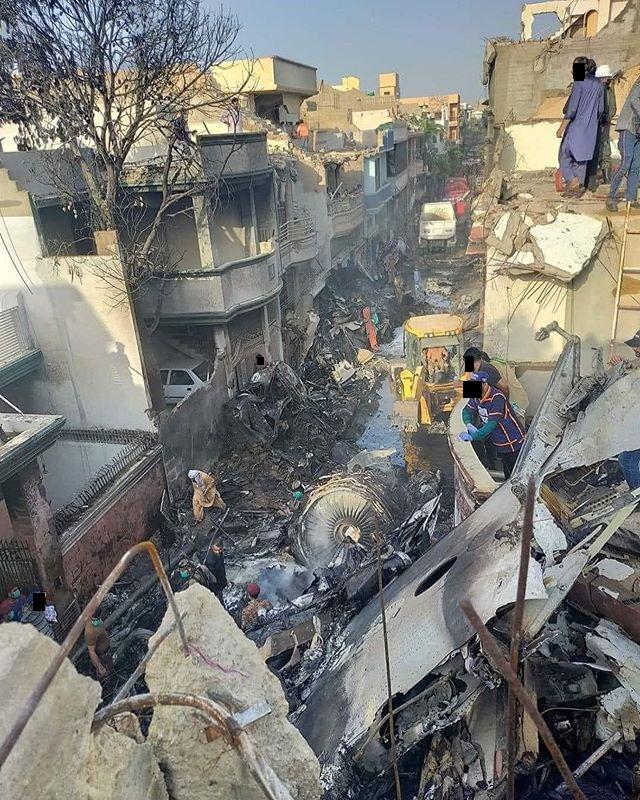}\par} &
``partially destroyed and surrounded by debris'', ``street littered with rubble'', ``smoke and dust'', ``overall atmosphere appears somber and tense'' &
destruction, accident, tension, debris, tragedy &
Negative/Negative &
``Debris/Destruction'', ``Morbid'', ``Accident'' \\

\hline

{\centering\includegraphics[width=1.5cm]{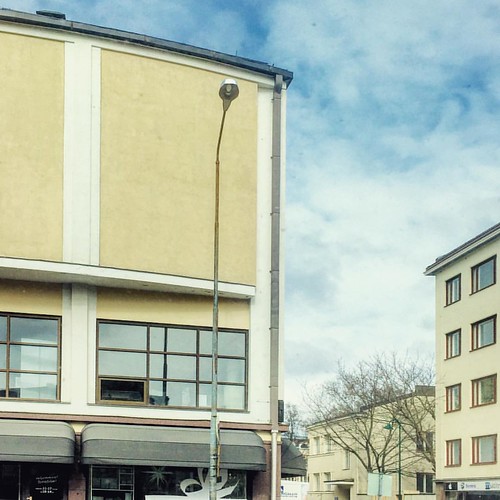}\par} &
``cloudy sky'', ``muted tone'', ``typical day in a city'', ``overall atmosphere is calm and somewhat quiet'' &
neutrality, ordinary urban setting, low emotional intensity &
Neutral/Neutral &
``Everyday image'', ``Meaningless'', ``It has positive and negative elements'' \\

\hline

\end{tabular}

\label{tab:interpretability_examples}

\end{adjustwidth}
\end{table*}

These findings inform research question \textbf {(Q4)}: traditional vision- and lexicon-based baselines remain competitive in simpler or coarse-grained conditions, where sentiment cues are visually salient, whereas the MLLM description-mediated approach, according to {\propAppCurto}, tends to yield larger benefits in settings where sentiment is tied to subtle, relational, or scene-level semantics. This helps clarify the conditions under which multimodal, reasoning-centric pipelines provide additional value for visual sentiment analysis.

\subsection*{Cross-dataset evaluation of {\propAppCurto}} 
\label{Sec:exp_vs_other_dataset}

In this section, we present a comparative evaluation of {\propAppCurto} on a distinct dataset focused exclusively on binary classification ($P_2$), i.e., positive versus negative sentiment, using the $\langle \sigma_3, P_2 \rangle$ and $\langle \sigma_5, P_2 \rangle$ settings. Importantly, here {\propAppCurto} is evaluated under a strict cross-dataset transfer protocol. The best-performing model trained on the source dataset (PerceptSent) is directly applied to the target dataset (DeepSent) without retraining, fine-tuning, parameter optimization, or prior exposure to target-domain images. This experimental setup allows us to assess the generalization and transfer capability of the proposed pipeline across a distinct image domain, in contrast to baseline methods originally trained and evaluated in-domain on DeepSent. To this end, {\propAppCurto} is fine-tuned using PerceptSent samples in a $P_2$ formulation, where original sentiment labels are restructured such that positive, slightly positive, and neutral are grouped as positive, while negative and slightly negative form the negative class. We report classification accuracy for consistency with prior work~\cite{you2015robust,outdoorsent2020}, which includes traditional hand-crafted features (GCH, LCH), shallow CNNs, and deep architectures such as VGG, Inception, ResNet, and DenseNet. Unlike these baselines, which are trained and validated directly on DeepSent using stratified $k$-fold cross-validation, our model is evaluated without exposure to the target dataset during training.

As reported in Table~\ref{tab:resultsSentVLMsTwitter}, although not trained or fine-tuned in DeepSent, {\propAppCurto} consistently outperforms all competing methods across both agreement levels ($\sigma_3$ and $\sigma_5$). Traditional hand-crafted approaches yield substantially lower performance, reaffirming the superiority of deep visual representations for sentiment analysis. Notably, {\propAppCurto} achieves absolute gains over the best-performing deep models reported by Oliveira~\etal~\cite{outdoorsent2020}, who have incorporated additional semantic features such as SUN scene descriptors and object-level cues from YOLO (e.g., guns, fire, homelessness) to improve sentiment prediction. In contrast, {\propAppCurto} attains higher accuracy without relying on such engineered features. This result suggests that the MLLM description-mediated pipeline captures semantic cues that transfer more readily across datasets, providing an indication of cross-dataset robustness. 

One possible explanation for this cross-dataset behavior is that the description-mediated pipeline partially reduces dependence on low-level visual characteristics by transforming images into semantic textual representations before sentiment classification. This intermediate representation may help the model transfer contextual and affective cues across datasets with different visual distributions. However, the results also suggest that robustness is still influenced by the nature of the target domain, particularly when sentiment depends on subtle or highly contextual cues. Therefore, although the proposed approach displays promising adaptability across datasets, its generalization ability remains tied to the quality and semantic consistency of the generated image descriptions.

\begin{table}[!htb]
\caption{Accuracy on the DeepSent dataset for two agreement levels: $\langle \sigma_3, P_2 \rangle$ (three-agree) and $\langle \sigma_5, P_2 \rangle$ (five-agree). Results include hand-crafted methods (GCH, LCH), shallow and deep CNN architectures, and the {\propAppCurto}, which is trained only on PerceptSent and evaluated on DeepSent without fine-tuning to assess cross-dataset generalization.}

\label{tab:resultsSentVLMsTwitter}
\footnotesize
\setlength{\tabcolsep}{10pt}
\renewcommand{\arraystretch}{1.2}
\centering
\begin{tabular}{cll}
\hline
\multirow{1}{*}{\textbf{Problem}} &  \multirow{1}{*}{\textbf{Method}} &  \multirow{1}{*}{\textbf{Accuracy}}  \\ \hline
\multirow{8}{*}{\normalsize $\langle \sigma_3$,$P_2 \rangle $} &  You~\etal~\cite{you2015robust} & $68.7$ \scriptsize $\pm 0$ \\
                                                    &  VGG16~\etal~\cite{outdoorsent2020} & $76.5$ \scriptsize $\pm 3.7$ \\
                                                    &  InceptionV3~\etal~\cite{outdoorsent2020} & $79.6$ \scriptsize $\pm 1.8$ \\
                                                    &  ResNet50~\etal~\cite{outdoorsent2020} & $81.5$ \scriptsize $\pm 1.0$ \\
                                                    &  DenseNet169~\etal~\cite{outdoorsent2020} & $81.4$ \scriptsize $\pm 1.4$ \\
                                                    &  GCH~\etal~\cite{you2015robust} & $66.0$ \scriptsize $\pm 0.0$ \\
                                                    &  LCH~\etal~\cite{you2015robust} & $66.4$  \scriptsize $\pm 0.0$ \\ &  Campos~\etal~\cite{campos2017pixels} & $74.9$  \scriptsize $\pm 0.04$ \\ 
                                                    &  \textbf{{\propAppCurto}} & \textbf{88.0} \scriptsize $\pm 0.7$ \\  \hline
\multirow{8}{*}{\normalsize $\langle \sigma_5$,$P_2 \rangle $} &  You~\etal~\cite{you2015robust} & $74.7$ \scriptsize $\pm 0$ \\
                                                    &  VGG16~\etal~\cite{outdoorsent2020} & $79.4$ \scriptsize $\pm 3.4$ \\
                                                    &  InceptionV3~\etal~\cite{outdoorsent2020} & $87.7$ \scriptsize $\pm 1.4$ \\
                                                    &  ResNet50~\etal~\cite{outdoorsent2020} & $86.5$ \scriptsize $\pm 3.1$ \\
                                                    &  DenseNet169~\etal~\cite{outdoorsent2020} & $88.3$ \scriptsize $\pm 1.6$ \\
                                                    &  GCH~\etal~\cite{you2015robust} & $68.4$ \scriptsize $\pm 0.0$ \\
                                                    &  LCH~\etal~\cite{you2015robust} & $71.0$  \scriptsize $\pm 0.0$ \\ &
                                                    Campos~\etal~\cite{campos2017pixels} & $83.0$  \scriptsize $\pm 0.03$ \\ 
                                                    &  \textbf{{\propAppCurto}}  & \textbf{95.6} \scriptsize $\pm 0.9$ \\  \hline
\end{tabular}
\end{table}

\begin{figure}[!htb]
\centering
\begin{tabular}{c c}
   \begin{minipage}[t]{0.3\textwidth}
    \includegraphics[width=3.0cm,height=2.5cm]{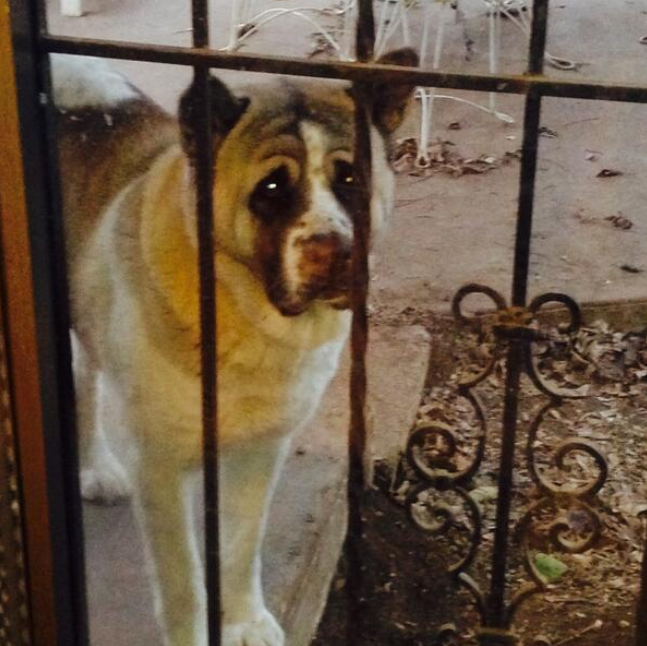} \vspace{-7pt}
    \begin{flushleft}
      \footnotesize
      Ground truth: Negative\\
      Campos~\etal~\cite{campos2017pixels}: Positive\\
      \textbf{\propAppCurto{}:} Negative
    \end{flushleft}
  \end{minipage} 
  &
  \begin{minipage}[t]{0.3\textwidth}
    \includegraphics[width=3.0cm,height=2.5cm]{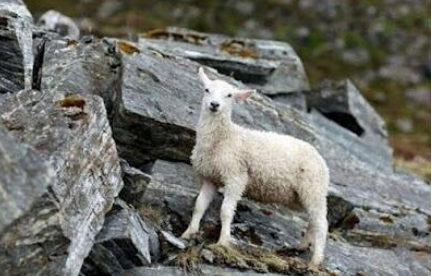} \vspace{-7pt}
    \begin{flushleft}
      \footnotesize
      Ground truth: Positive\\
      Campos~\etal~\cite{campos2017pixels}: Negative\\
      \textbf{\propAppCurto{}:} Positive
    \end{flushleft}
  \end{minipage}\\ \vspace{5pt}
  \footnotesize (a) & \footnotesize (b)\\
  \begin{minipage}[t]{0.3\textwidth}
    \includegraphics[width=3.0cm,height=2.5cm]{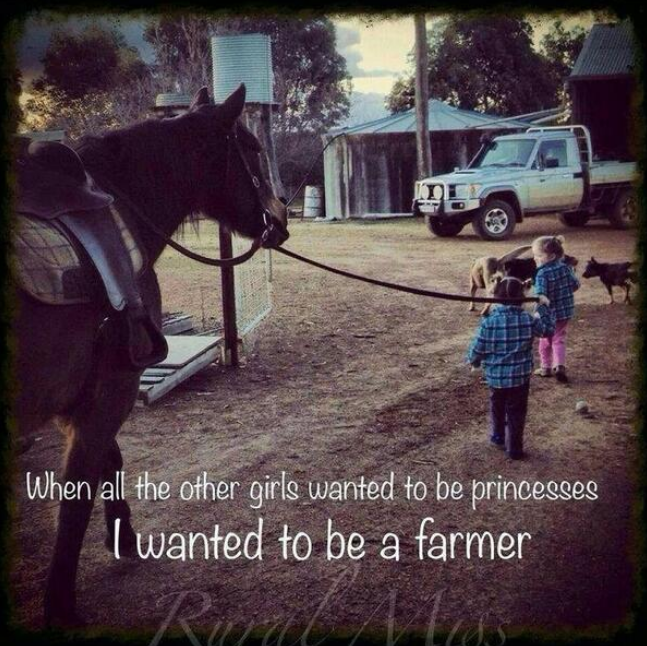} \vspace{-7pt} \vspace{-2pt}
    \begin{flushleft}
      \footnotesize
      Ground truth: Positive\\ 
      Campos~\etal~\cite{campos2017pixels}: Negative\\ 
      \textbf{\propAppCurto{}:} Positive
    \end{flushleft}
  \end{minipage}
  &
  \begin{minipage}[t]{0.3\textwidth}
    \includegraphics[width=3.0cm,height=2.5cm]{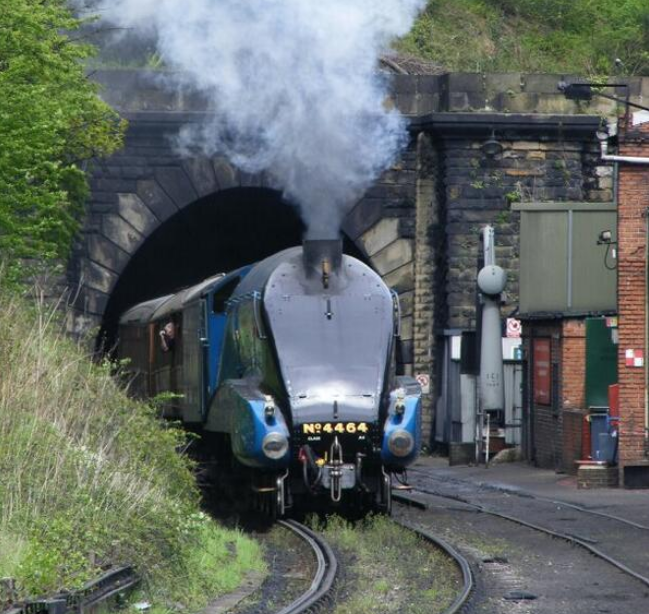} \vspace{-7pt} 
    \begin{flushleft}
      \footnotesize
      Ground truth: Positive\\ 
      Oliveira~\etal~\cite{outdoorsent2020}: Negative\\ 
      \textbf{\propAppCurto{}:} Negative 
    \end{flushleft}
  \end{minipage}\\ \vspace{5pt}
  \footnotesize (c) & \footnotesize (d) 
\end{tabular}
\caption{Qualitative comparison of sentiment predictions on DeepSent image samples. This figure is original to this work. Images (a-c) are derived from the DeepSent GitHub repository~\cite{CamposGit}, distributed under the MIT License; Image (d) is derived from the PerceptSent dataset~\cite{Cesar}, distributed under CC BY 4.0 license.}
\label{fig:qualitative_twitter}
\end{figure}

 Fig.~\ref{fig:qualitative_twitter} presents selected images from the DeepSent dataset that are misclassified by the methods of Oliveira~\etal~\cite{outdoorsent2020} and Campos~\etal~\cite{campos2017pixels}, along with the corresponding outputs from {\propAppCurto}. These examples are used to illustrate how the MLLM description-mediated pipeline behaves in cases where visual-only models struggle. In Figs.~\ref{fig:qualitative_twitter}(a,b), the MLLM {\imgDescrip}s correctly identify animals and contextual elements related to rural or mountainous landscapes. In Fig.~\ref{fig:qualitative_twitter}(a), however, the negative classification may be attributed to descriptions such as ``\textit{a large dog standing outside, looking through a metal gate}" and ``\textit{a mix of dirt and possibly some fallen leaves}." These examples suggest that certain description fragments may introduce affective cues that influence the downstream sentiment prediction. It is worth noting that not all descriptive sentences convey exclusively positive or negative semantics, which reinforces the challenge of disentangling scene description from affect interpretation in this pipeline. 

In Fig.~\ref{fig:qualitative_twitter}(d), our model also produces an incorrect outcome. We conjecture that the description ``\textit{thick white steam is billowing from the smokestack}'' generated by the MLLM may have influenced the negative sentiment classification --- potentially outweighing the more neutral or positive phrase ``\textit{the overall scene captures a blend of industrial heritage and natural beauty}," despite the image likely depicting a tourism-related train. This example highlights a broader limitation of the MLLM description-mediated pipeline for visual sentiment analysis: when neutral contextual information coexists with localized cues that may carry negative affective associations, the final prediction can become overly sensitive to how such elements are verbalized in the generated descriptions.

\subsection*{Limitations}

A limitation of the proposed approach is its dependence on the quality and consistency of the captions generated by the MLLMs, since inaccurate (noisy), incomplete (omitted contextual cues), or hallucinated descriptions may propagate errors to the downstream sentiment classification framework and contribute to some of the observed misclassifications. More generally, the effectiveness of the description-mediated pipeline strongly depends on how MLLMs verbalize subtle contextual and affective cues, as small differences in emphasis or wording may substantially influence the final sentiment prediction, particularly in ambiguous or context-dependent scenes. In addition, image sentiment analysis is inherently subjective, and sentiment labels may reflect annotator interpretation and cultural bias, which can affect both training and evaluation. 
Recent work has explored persona-based prompting as a way to account for such variation in urban sentiment perception, but also shows that persona conditioning may produce only limited behavioral differentiation and may not fully capture fine-grained human perceptual judgments \cite{urbcom26-neemias}. Although this motivates further investigation of persona-aware sentiment analysis, persona conditioning is outside the scope of the present study. Moreover, although the framework was evaluated using both open-access and proprietary MLLMs, part of the experimental analysis relies on a proprietary model (GPT/OAI), which may limit full reproducibility due to possible future changes in model availability or behavior.

Another limitation concerns the scope of the benchmark comparison. While prior multimodal sentiment analysis works have explored specialized fusion-based architectures, many rely on jointly trained visual-textual pipelines and task-specific optimization strategies, often requiring dedicated training procedures and substantial computational resources. In contrast, the goal of this work is not to exhaustively benchmark all multimodal sentiment architectures, but rather to systematically analyze MLLM-based visual reasoning and description-mediated pipelines under a unified and reproducible evaluation framework. Future work may explore direct comparisons between description-mediated MLLM pipelines and end-to-end multimodal fusion architectures specifically designed for affective computing tasks.

\section*{Conclusions and future work}~\label{sec:conclusions}

This study investigated how Multimodal Large Language Models (MLLMs) can support visual sentiment analysis by operating within a two-stage description-mediated pipeline. Through a systematic set of experiments, we evaluated this pipeline across multiple configurations and addressed the four research questions posed in the study. Concerning Q1, we found that direct image-based sentiment classification using MLLMs (\TaskOne) can be effective, particularly when employing higher-capacity models such as {\VLMoai}, as instantiated in our evaluation, which consistently outperformed locally run open-weights alternatives. At the same time, the results indicate that observed differences across models reflect a combination of capacity, architectural design, and training data characteristics, rather than scale alone.

Regarding Q2, transforming images into textual descriptions and classifying them using pre-trained LLMs (\TaskTwoA) yielded moderate but consistent gains for ModernBERT in most scenarios. Addressing Q3, we observed that fine-tuning LLMs on MLLM-generated descriptions (\TaskTwoB) consistently led to relevant performance improvements across all MLLM+LLM combinations and label configurations. These findings suggest that effective visual sentiment reasoning depends not only on rich visual grounding but also on adapting the language-reasoning component to the target task. Notably, ModernBERT (395M parameters), when fine-tuned on MLLM-generated descriptions, achieved strong results in high-agreement settings, surpassing larger models such as LLAMA-3 (8B parameters) under high-agreement conditions. This highlights that carefully adapted mid-sized models can compete with substantially larger ones when supported by informative description representations.

Finally, in response to Q4, \propAppCurto{}, representing the best combination tested of the MLLM description-mediated pipeline, performed better than visual-only baselines (e.g., CNNs and Swin Transformers) as well as rule-based text-only classifiers (e.g., VADER), and exhibited encouraging cross-dataset robustness in our evaluation in settings where sentiment depends on transferable semantic cues.

Rather than focusing the analysis on a purely performance-driven alternative, we view a MLLM description-mediated pipeline for visual sentiment analysis also as a useful analytical tool. The textual descriptions produced by MLLMs provide an interpretable intermediate layer that helps clarify the basis of model decisions by externalizing perceptual and contextual cues, which is particularly valuable in the analysis of user-generated content, where transparency and explainability support trust and qualitative assessment.

Although we evaluated GPT-4o mini, MiniGPT-4, and DeepSeek-VL2-Tiny, our methodology is model-agnostic. The consistent behavior observed across different architectures suggests that the conclusions are not limited to the specific models tested and that future MLLMs may further improve results while preserving interpretability benefits. Moreover, despite GPT-4o mini achieving the best overall performance, open and open-weights alternatives (\VLMds, \VLMos) remain attractive due to accessibility and deployment flexibility. Exploring larger open-weights variants and additional multimodal model families, therefore, represents a natural direction for follow-up work.

One of the study's biggest contributions lies in demonstrating how an image description-mediated pipeline, leveraged by MLLMs, can structure sentiment inference, helping to decouple visual scene interpretation from sentiment reasoning over generated descriptions. This provides a practical bridge between performance and interpretability, offering a traceable path from images to predictions, which is particularly relevant for domains where transparency and accountability are desirable, such as healthcare, policy analysis, and online content moderation.

Future research may extend this work by exploring different hyperparameter configurations, and conducting more systematic analyses of failure modes to better understand the performance and stability of the proposed approach. Studies involving larger and more diverse datasets may also help improve the robustness and generalization of description-mediated sentiment reasoning across different visual domains. Another promising direction is the comparison and integration of description-mediated MLLM pipelines with multimodal fusion strategies that jointly model visual and textual representations, aiming to better understand their complementary strengths in visual sentiment analysis.
Future studies should also examine the computational efficiency of MLLM- and LLM-based sentiment analysis pipelines in production settings, including runtime performance, latency, resource consumption, and the trade-offs between predictive accuracy and computational cost. 

Additionally, future work could involve the development of iterative or feedback-driven refinement strategies for MLLM-generated descriptions, particularly in cases where important contextual or affective cues are omitted. In this context, interactive annotation workflows in which users refine ambiguous or incomplete descriptions may improve both dataset quality and downstream model reliability. Complementarily, explainable artificial intelligence techniques could be incorporated to better associate visual regions and linguistic cues with sentiment predictions, improving the interpretability of the reasoning process.

\subsection*{Ethics statement}
\label{sec:ehtics}

This study uses the previously published PerceptSent and DeepSent datasets; therefore, this work constitutes a secondary analysis of existing publicly available, anonymized data. No new data were collected from human participants, no direct interaction with participants occurred, no identifiable personal information was accessed, and the study does not involve minors. Because only secondary analysis of public anonymized datasets was performed, informed consent for this study was not applicable.

We emphasize that the proposed models are intended for research purposes and that automated sentiment inference from images may be sensitive to contextual and subjective interpretation biases.

\bibliographystyle{plos2025}

\begin{thebibliography}{10}

\bibitem{VLMforSentAnalysis2023}
Bustos C, Civit C, Du B, Solé-Ribalta A, Lapedriza A.
\newblock On the use of Vision-Language models for Visual Sentiment Analysis: a study on {CLIP}.
\newblock In: Proc. of ACII; 2023. p. 1-8.
\newblock \href {http://dx.doi.org/10.1109/ACII59096.2023.10388075} {doi:10.1109/ACII59096.2023.10388075}.

\bibitem{zisad2021integrated}
Zisad SN, Chowdhury E, Hossain MS, Islam RU, Andersson K.
\newblock An Integrated Deep Learning and Belief Rule-Based Expert System for Visual Sentiment Analysis under Uncertainty.
\newblock Algorithms. 2021;14(7).
\newblock \href {http://dx.doi.org/10.3390/a14070213} {doi:10.3390/a14070213}.

\bibitem{chandrasekaran2022visual}
Chandrasekaran G, Antoanela N, Andrei G, Monica C, Hemanth J.
\newblock Visual Sentiment Analysis Using Deep Learning Models with Social Media Data.
\newblock Applied Sciences. 2022;12(3).
\newblock \href {http://dx.doi.org/10.3390/app12031030} {doi:10.3390/app12031030}.

\bibitem{ortis2020survey}
Ortis A, Farinella GM, Battiato S.
\newblock Survey on visual sentiment analysis.
\newblock IET Image Processing. 2020;14(8):1440-56.

\bibitem{PerceptSent}
Lopes CR, Minetto R, Delgado MR, Silva TH.
\newblock {P}ercept{S}ent - Exploring Subjectivity in a Novel Dataset for Visual Sentiment Analysis.
\newblock IEEE Transactions on Affective Computing. 2023;14(3).
\newblock \href {http://dx.doi.org/10.1109/TAFFC.2022.3225238} {doi:10.1109/TAFFC.2022.3225238}.

\bibitem{kil2024mllm}
Kil J, Mai Z, Lee J, Chowdhury A, Wang Z, Cheng K, et~al.
\newblock {MLLM}-{C}omp{B}ench: A comparative reasoning benchmark for multimodal {LLM}s.
\newblock Advances in Neural Information Processing Systems. 2024;37:28798-827.

\bibitem{zhang2024mm}
Zhang D, Yu Y, Dong J, Li C, Su D, Chu C, et~al.
\newblock {MM-LLM}s: Recent advances in multimodal large language models.
\newblock arXiv preprint arXiv:240113601. 2024.

\bibitem{cheng2025evaluating}
Cheng Z, Xu B, Gong L, Song Z, Zhou T, Zhong S, et~al.
\newblock Evaluating {MLLM}s with multimodal multi-image reasoning benchmark.
\newblock arXiv preprint arXiv:250604280. 2025.

\bibitem{10.1145/3746027.3755591}
Luo M, Jiang Y, Mai S.
\newblock Towards Explainable Fusion and Balanced Learning in Multimodal Sentiment Analysis.
\newblock In: Proceedings of the 33rd ACM International Conference on Multimedia. MM '25. New York, NY, USA: Association for Computing Machinery; 2025. p. 1997–2006.
\newblock Available from: \url{https://doi.org/10.1145/3746027.3755591}. \href {http://dx.doi.org/10.1145/3746027.3755591} {doi:10.1145/3746027.3755591}.

\bibitem{ansari2025role}
Ansari ZA, Tripathi MM, Ahmed R.
\newblock The role of explainable AI in enhancing breast cancer diagnosis using machine learning and deep learning models.
\newblock Discover Artificial Intelligence. 2025;5(1):75.

\bibitem{10.1145/3723005}
Zhang X, Zhang T, Sun L, Zhao J, Jin Q.
\newblock Exploring Interpretability in Deep Learning for Affective Computing: A Comprehensive Review.
\newblock ACM Trans Multimedia Comput Commun Appl. 2025 Jul;21(7).
\newblock Available from: \url{https://doi.org/10.1145/3723005}. \href {http://dx.doi.org/10.1145/3723005} {doi:10.1145/3723005}.

\bibitem{ZHANG2025112812}
Zhang M, Pan Y, Cui Q, Lü Y, Yu W.
\newblock Multimodal LLM for enhanced Alzheimer’s Disease diagnosis: Interpretable feature extraction from Mini-Mental State Examination data.
\newblock Experimental Gerontology. 2025;208:112812.
\newblock Available from: \url{https://www.sciencedirect.com/science/article/pii/S053155652500141X}. \href {http://dx.doi.org/https://doi.org/10.1016/j.exger.2025.112812} {doi:https://doi.org/10.1016/j.exger.2025.112812}.

\bibitem{ansari2025optimized}
Ansari ZA, Ansari MSH, Khan A, Jhanjhi N, Rathnamma G.
\newblock Optimized and interpretable machine learning framework for early breast cancer detection.
\newblock Health and Technology. 2025;15(6):1135-47.

\bibitem{zhu2024minigpt}
Zhu D, Chen J, Shen X, Li X, Elhoseiny M.
\newblock Mini{GPT}-4: Enhancing Vision-Language Understanding with Advanced Large Language Models.
\newblock In: Proc. of ICLR; 2024. .

\bibitem{openai2024gpt4ocard}
OpenAI, :, Hurst A, Lerer A, Goucher AP, Perelman A, et~al.
\newblock {GPT-4o} System Card.
\newblock arXiv preprint arXiv:241021276. 2024.

\bibitem{wu2024deepseek}
Wu Z, Chen X, Pan Z, Liu X, Liu W, Dai D, et~al.
\newblock {D}eep{S}eek-{VL2}: Mixture-of-experts vision-language models for advanced multimodal understanding.
\newblock arXiv preprint arXiv:241210302. 2024.

\bibitem{lewis2020bart}
Lewis M, Liu Y, Goyal N, Ghazvininejad M, Mohamed A, Levy O, et~al.
\newblock {BART}: Denoising Sequence-to-Sequence Pre-training for Natural Language Generation, Translation, and Comprehension.
\newblock In: Proc. of ACL; 2020. .

\bibitem{williams2018broad}
Williams A, Nangia N, Bowman SR.
\newblock A Broad-Coverage Challenge Corpus for Sentence Understanding through Inference.
\newblock In: Proceedings of NAACL-HLT; 2018. p. 1112-22.

\bibitem{warner2024modernBert}
Warner B, et~al.
\newblock Smarter, Better, Faster, Longer: A Modern Bidirectional Encoder for Fast, Memory Efficient, and Long Context Finetuning and Inference.
\newblock In: Proc. of ACL; 2025. .

\bibitem{dubey2024llama3}
Grattafiori A, Dubey A, Jauhri A, Pandey A, Kadian A, Al-Dahle A, et~al.
\newblock The {L}lama 3 herd of models.
\newblock arXiv preprint arXiv:240721783. 2024.

\bibitem{jurek2015improved}
Jurek A, Mulvenna MD, Bi Y.
\newblock Improved lexicon-based sentiment analysis for social media analytics.
\newblock Security Informatics. 2015;4:1-13.

\bibitem{jha2018novel}
Jha V, Savitha R, Shenoy PD, Venugopal K, Sangaiah AK.
\newblock A novel sentiment aware dictionary for multi-domain sentiment classification.
\newblock Computers \& Electrical Engineering. 2018;69.

\bibitem{bernabe2020context}
Bernab{\'e}-Moreno J, Tejeda-Lorente A, Herce-Zelaya J, Porcel C, Herrera-Viedma E.
\newblock A context-aware embeddings supported method to extract a fuzzy sentiment polarity dictionary.
\newblock Knowledge-Based Systems. 2020;190:105236.

\bibitem{viegas2020exploiting}
Viegas F, Alvim MS, Canuto S, Rosa T, Gon{\c{c}}alves MA, Rocha L.
\newblock Exploiting semantic relationships for unsupervised expansion of sentiment lexicons.
\newblock Information Systems. 2020;94:101606.

\bibitem{10478509}
He A, Abisado M.
\newblock Text Sentiment Analysis of {D}ouban Film Short Comments Based on {BERT}-{CNN}-{BiLSTM}-{A}tt Model.
\newblock IEEE Access. 2024;12:45229-37.
\newblock \href {http://dx.doi.org/10.1109/ACCESS.2024.3381515} {doi:10.1109/ACCESS.2024.3381515}.

\bibitem{LIANG2022107643}
Liang B, Su H, Gui L, Cambria E, Xu R.
\newblock Aspect-based sentiment analysis via affective knowledge enhanced graph convolutional networks.
\newblock Knowledge-Based Systems. 2022;235:107643.
\newblock \href {http://dx.doi.org/https://doi.org/10.1016/j.knosys.2021.107643} {doi:https://doi.org/10.1016/j.knosys.2021.107643}.

\bibitem{WU2022107736}
Wu H, Zhang Z, Shi S, Wu Q, Song H.
\newblock Phrase dependency relational graph attention network for Aspect-based Sentiment Analysis.
\newblock Knowledge-Based Systems. 2022;236:107736.
\newblock \href {http://dx.doi.org/https://doi.org/10.1016/j.knosys.2021.107736} {doi:https://doi.org/10.1016/j.knosys.2021.107736}.

\bibitem{Yin_Zhong_2024}
Yin S, Zhong G.
\newblock {T}ext{GT}: A Double-View Graph Transformer on Text for Aspect-Based Sentiment Analysis.
\newblock AAAI Conference on Artificial Intelligence. 2024 Mar;38(17):19404-12.
\newblock \href {http://dx.doi.org/10.1609/aaai.v38i17.29911} {doi:10.1609/aaai.v38i17.29911}.

\bibitem{stappen2021sentiment}
Stappen L, Baird A, Cambria E, Schuller BW.
\newblock Sentiment analysis and topic recognition in video transcriptions.
\newblock IEEE Intelligent Systems. 2021;36(2):88-95.

\bibitem{Ji2016}
Ji R, Cao D, Zhou Y, Chen F.
\newblock Survey of visual sentiment prediction for social media analysis.
\newblock Frontiers of Computer Science. 2016 Aug;10(4):602-11.

\bibitem{you2015robust}
You Q, Luo J, Jin H, Yang J.
\newblock Robust Image Sentiment Analysis Using Progressively Trained and Domain Transferred Deep Networks.
\newblock In: Proc.\ of AAAI. AAAI Press; 2015. p. 381-8.

\bibitem{cnn1}
Chen T, Borth D, Darrell T, Chang S.
\newblock {D}eep{S}enti{B}ank: Visual Sentiment Concept Classification with Deep Convolutional Neural Networks.
\newblock arXiv preprint arXiv:14108586. 2014.

\bibitem{SONG2018218}
Song K, Yao T, Ling Q, Mei T.
\newblock Boosting image sentiment analysis with visual attention.
\newblock Neurocomputing. 2018;312.
\newblock \href {http://dx.doi.org/https://doi.org/10.1016/j.neucom.2018.05.104} {doi:https://doi.org/10.1016/j.neucom.2018.05.104}.

\bibitem{outdoorsent2020}
Oliveira WBd, Dorini LB, Minetto R, Silva TH.
\newblock Outdoor{S}ent: Sentiment Analysis of Urban Outdoor Images by Using Semantic and Deep Features.
\newblock ACM Trans on Information Systems. 2020;38(3).
\newblock \href {http://dx.doi.org/10.1145/3385186} {doi:10.1145/3385186}.

\bibitem{PANDEY2024111206}
Pandey A, Vishwakarma DK.
\newblock Progress, achievements, and challenges in multimodal sentiment analysis using deep learning: A survey.
\newblock Applied Soft Computing. 2024;152:111206.
\newblock \href {http://dx.doi.org/https://doi.org/10.1016/j.asoc.2023.111206} {doi:https://doi.org/10.1016/j.asoc.2023.111206}.

\bibitem{das2023image}
Das R, Singh TD.
\newblock Image--text multimodal sentiment analysis framework of assamese news articles using late fusion.
\newblock ACM Transactions on Asian and Low-Resource Language Information Processing. 2023;22(6):1-30.

\bibitem{zhu2022multimodal}
Zhu T, Li L, Yang J, Zhao S, Liu H, Qian J.
\newblock Multimodal sentiment analysis with image-text interaction network.
\newblock IEEE transactions on multimedia. 2022;25:3375-85.

\bibitem{pan2024multi}
Pan J, Lu J, Wang S.
\newblock A multi-stage visual perception approach for image emotion analysis.
\newblock IEEE Trans on Affective Computing. 2024.

\bibitem{wang2023dual}
Wang D, Tian C, Liang X, Zhao L, He L, Wang Q.
\newblock Dual-perspective fusion network for aspect-based multimodal sentiment analysis.
\newblock IEEE Transactions on Multimedia. 2023;26:4028-38.

\bibitem{liu2025unveiling}
Liu Y, Liang Z, Wang Y, Wu X, Tang F, He M, et~al.
\newblock Unveiling the Ignorance of {MLLM}s: Seeing Clearly, Answering Incorrectly.
\newblock In: Proc. of CVPR; 2025. p. 9087-97.

\bibitem{zhao2024exploring}
Zhao S, Tuan LA, Fu J, Wen J, Luo W.
\newblock Exploring clean label backdoor attacks and defense in language models.
\newblock IEEE/ACM Transactions on Audio, Speech, and Language Processing. 2024.

\bibitem{zhang2023sentimentanalysiseralarge}
Zhang W, Deng Y, Liu B, Pan S, Bing L.
\newblock Sentiment Analysis in the Era of Large Language Models: A Reality Check.
\newblock In: Proc. of NAACL; 2024. .

\bibitem{villalobos2024position}
Villalobos P, Ho A, Sevilla J, Besiroglu T, Heim L, Hobbhahn M.
\newblock Position: Will we run out of data? {L}imits of {LLM} scaling based on human-generated data.
\newblock In: Proc. of ICML; 2024. .

\bibitem{hong2024only}
Hong H, Wang S, Huang Z, Wu Q, Liu J.
\newblock Why only text: empowering vision-and-language navigation with multi-modal prompts.
\newblock In: Proc. of IJCAI; 2024. .

\bibitem{lu2024deepseek}
Lu H, Liu W, Zhang B, Wang B, Dong K, Liu B, et~al.
\newblock {D}eep{S}eek-{VL}: towards real-world vision-language understanding.
\newblock arXiv preprint arXiv:240305525. 2024.

\bibitem{xiao2025exploring}
Xiao L, Mao R, Zhao S, Lin Q, Jia Y, He L, et~al.
\newblock Exploring Cognitive and Aesthetic Causality for Multimodal Aspect-Based Sentiment Analysis.
\newblock IEEE Transactions on Affective Computing. 2025.

\bibitem{huang2023language}
Huang S, Dong L, Wang W, Hao Y, Singhal S, Ma S, et~al.
\newblock Language is not all you need: Aligning perception with language models.
\newblock Advances in Neural Information Processing Systems. 2023;36:72096-109.

\bibitem{Cesar}
Lopes CR, Minetto R, Delgado M, Silva TH. {PerceptSent} dataset. GitHub; 2022.
\newblock \url{https://github.com/ceslop84/perceptsent}.

\bibitem{CamposGit}
Campos. From Pixels to Sentiment: Fine-tuning CNNs for Visual Sentiment Prediction. GitHub; 2017.
\newblock \url{https://github.com/imatge-upc/sentiment-2017-imavis}.

\bibitem{zheng2023judgingllmasajudgemtbenchchatbot}
Zheng L, Chiang WL, Sheng Y, Zhuang S, Wu Z, Zhuang Y, et~al.
\newblock Judging {LLM}-as-a-judge with {MT}-bench and Chatbot Arena.
\newblock In: Proc. of NIPS; 2023. .

\bibitem{touvron2023llama}
Touvron H, Lavril T, Izacard G, Martinet X, Lachaux MA, Lacroix T, et~al.
\newblock {LLaMA}: Open and efficient foundation language models.
\newblock arXiv preprint arXiv:230213971. 2023.

\bibitem{li2023blip}
Li J, Li D, Savarese S, Hoi S.
\newblock {BLIP}-2: Bootstrapping language-image pre-training with frozen image encoders and large language models.
\newblock In: Proc. of ICML; 2023. .

\bibitem{fang2023eva}
Fang Y, Wang W, Xie B, Sun Q, Wu L, Wang X, et~al.
\newblock {EVA}: Exploring the limits of masked visual representation learning at scale.
\newblock In: IEEE CVPR; 2023. .

\bibitem{hutto2014vader}
Hutto C, Gilbert E.
\newblock {VADER}: A parsimonious rule-based model for sentiment analysis of social media text.
\newblock In: Proc. of ICWSM. vol.~8; 2014. .

\bibitem{campos2017pixels}
Campos V, Jou B, i~Nieto XG.
\newblock From pixels to sentiment: {F}ine-tuning {CNN}s for visual sentiment prediction.
\newblock Image and Vision Computing. 2017;65:15  22.
\newblock \href {http://dx.doi.org/https://doi.org/10.1016/j.imavis.2017.01.011} {doi:https://doi.org/10.1016/j.imavis.2017.01.011}.

\bibitem{urbcom26-neemias}
da~Silva NB, Minetto R, Silver D, Silva TH.
\newblock Stable Behavior, Limited Variation: Persona Validity in LLM Agents for Urban Sentiment Perception.
\newblock In: Proc. of IEEE DCOSS-IoT-UrbCom. Reykjavik, Iceland; 2026. .

\end{thebibliography}

\nolinenumbers

%
%

\begin{flushleft}
{\Large
\textbf{Supplementary Information} 
}
\newline
\\

%
%






\end{flushleft}


%


\section{Appendix - Code and Data Availability}

All source code, experimental scripts, trained models, evaluation routines, and generated intermediate descriptions used in this study are publicly available at:

\begin{center}
\texttt{\url {https://github.com/neemiasbsilva/multimodal-LLMs-see-sentiment}}
\end{center}

The repository includes: (i) scripts for data preprocessing; (ii) training and evaluation pipelines; (iii) prompts used in all experiments; (iv) generated image descriptions produced by MLLMs; (v) fine-tuned model checkpoints; (vi) experimental configuration files.

Releasing the generated descriptions and experimental artifacts helps mitigate reproducibility limitations associated with proprietary API-based models.

\section{Appendix - Prompt Templates}

\subsection{Task 1 -- Direct Image Sentiment Classification}

The following prompt template was used for direct sentiment classification from images:

\begin{quote}
\texttt{Analyze this image, and classify it as \{L\} sentiments, do not describe the image, and select only one class.}
\end{quote}

where \texttt{\{L\}} corresponds to the sentiment labels associated with each problem formulation:
\begin{itemize}
    \item \textbf{P2}: \{positive, negative\}
    \item \textbf{P3}: \{positive, neutral, negative\}
    \item \textbf{P5}: \{positive, slightly positive, neutral, slightly negative, negative\}
\end{itemize}

\subsection{Task 2 -- Image Description Generation}

The following prompt template was used to generate image descriptions:

\begin{quote}
\texttt{Describe this image in details.}
\end{quote}

The generated descriptions were subsequently used as input to text-only LLMs for sentiment classification.

\subsection{Task 2 -- Text-based Sentiment Classification}

The following prompt template was used for text-based sentiment classification:

\begin{quote}
\texttt{What is the sentiment of this description? Please choose an answer from \{LABEL\_SET\}.}
\end{quote}

Example for P3:

\begin{quote}
\texttt{What is the sentiment of this description? Please choose an answer from \{"Positive": 2, "Negative": 0, "Neutral": 1\}.}
\end{quote}

\section{Appendix - Extended Benchmark with Additional Recent MLLMs}
\label{appendix:extendedbenchmark}

To further investigate whether the trends observed in the main experiments generalize across recent multimodal large language models (MLLMs), we conducted an additional benchmark using three newer MLLMs: Gemini, Phi-4, and Gemma-4-E4B. These models were evaluated uder two distinct paradigms: (i) direct image sentiment classification (Task~1), and sentiment classification using MLLM-generated descriptions with a fine-tuned MBERT classifier (Task~2$_b$) (our best setup result for {\propAppCurto} framework).

To optimize computational efficiency without compromising analytical rigor, we restricted this supplementary evaluation to the most robust configuration identified in our primary analysis: the MBERT-based pipeline ({\propAppCurto}). GPT (OAI), the best-performing model from the main experiments, is included as a reference baseline with focus on the best setup applying finetuning.

The outcomes of this extended analysis, presented in Table~\ref{tab:extendedMLLMbenchmark}, confirm that the fundamental dynamics observed in the main study remain remarkably consistent across these newer MLLM families. In particular, description-mediated reasoning combined with
fine-tuned MBERT (Task~2$_b$) consistently achieves the strongest overall performance, while direct image classification (Task~1) remains more sensitive to
class granularity and annotator agreement. Although Gemini achieves competitive results in direct classification, the GPT (OAI)+MBERT configuration continues to provide the best overall performance across the evaluated settings.

\begin{table}[!htb]
\begin{adjustwidth}{-2.25in}{0in}
  \caption{
Weighted $F$-score (\%, $\pm95\%$ CI, 5-fold CV) for Task~1 (direct sentiment classification)
and Task~$2_b$
(MLLM-generated descriptions classified using fine-tuned MBERT)
across additional recent MLLMs and the best-performing existing
model (GPT (OAI)$^\dagger$, from Table~1).
Bold values indicate the best results.
}
    \centering
    \resizebox{\linewidth}{!}{%
    \begin{tabular}{c|cccc|cccc}
        \hline
\multirow{2}{*}{\textbf{Problem}}
  & \multicolumn{4}{c|}{\textbf{Task 1 (Direct Classification)}}
 & \multicolumn{4}{c}{\textbf{Task $2_b$ (MBERT fine-tuned)}}
\\

  & \multicolumn{1}{c}{GPT (OAI)$^\dagger$}
  & \multicolumn{1}{c}{Gemini}
  & \multicolumn{1}{c}{Phi-4}
  & \multicolumn{1}{c|}{Gemma-4-E4B}
  & \multicolumn{1}{c}{GPT (OAI) $^\dagger$}
  & \multicolumn{1}{c}{Gemini}
  & \multicolumn{1}{c}{Phi-4}
  & \multicolumn{1}{c}{Gemma-4-E4B}
\\
\hline
$\langle\sigma_3,P_5\rangle$
  & $44.5$ \scriptsize $\pm 3.1$
  & $51.3$ \scriptsize $\pm 1.5$
  & $42.2$ \scriptsize $\pm 3.7$
  & $46.8$ \scriptsize $\pm 4.8$
  & \textbf{58.4} \scriptsize $\pm 4.0$
  & $56.9$ \scriptsize $\pm 1.3$
  & $53.2$ \scriptsize $\pm 2.8$
  & $57.4$ \scriptsize $\pm 2.8$
\\\hline
$\langle\sigma_3,P_3\rangle$
  & $61.2$ \scriptsize $\pm 2.9$
  & $69.8$ \scriptsize $\pm 1.3$
  & $61.2$ \scriptsize $\pm 2.9$
  & $60.7$ \scriptsize $\pm 2.8$
  & \textbf{77.5} \scriptsize $\pm 0.9$
  & $76.3$ \scriptsize $\pm 3.0$
  & $73.1$ \scriptsize $\pm 2.7$
  & $75.4$ \scriptsize $\pm 1.2$
\\\hline
$\langle\sigma_5,P_5\rangle$
  & $75.8$ \scriptsize $\pm 4.7$
  & $78.5$ \scriptsize $\pm 6.0$
  & $78.1$ \scriptsize $\pm 6.7$
  & $78.3$ \scriptsize $\pm 4.2$
  & \textbf{84.4} \scriptsize $\pm 4.2$
  & $81.4$ \scriptsize $\pm 6.6$
  & $79.0$ \scriptsize $\pm 5.5$
  & $83.3$ \scriptsize $\pm 4.8$
\\\hline
$\langle\sigma_5,P_3\rangle$
  & $87.7$ \scriptsize $\pm 1.8$
  & $88.2$ \scriptsize $\pm 2.2$
  & $82.1$ \scriptsize $\pm 1.5$
  & $81.0$ \scriptsize $\pm 1.6$
  & \textbf{95.8} \scriptsize $\pm 0.9$
  & $94.5$ \scriptsize $\pm 2.1$
  & $93.3$ \scriptsize $\pm 2.1$
  & $94.2$ \scriptsize $\pm 0.7$
\\\hline
    \end{tabular}
    }
    \label{tab:extendedMLLMbenchmark}
    \end{adjustwidth}
\end{table}


\end{document}